\documentclass[journal]{IEEEtran}
\pdfoutput=1
\usepackage{cite}

\ifCLASSINFOpdf
  \usepackage[pdftex]{graphicx}
\else
\fi
\usepackage{amsmath}
\usepackage{amssymb}
\usepackage{array}

\ifCLASSOPTIONcompsoc
  \usepackage[caption=false,font=normalsize,labelfont=sf,textfont=sf]{subfig}
\else
  \usepackage[caption=false,font=footnotesize]{subfig}
\fi

\usepackage{stfloats}
\usepackage{url}

\usepackage{xspace}
\newcommand*{\eg}{\emph{e.g.,}\@\xspace}
\newcommand*{\ie}{\emph{i.e.,}\@\xspace}

\makeatletter
\newcommand*{\etc}{%
    \@ifnextchar{.}%
        {etc}%
        {etc.\@\xspace}%
}
\makeatother

\makeatletter
\newcommand*{\etal}{%
    \@ifnextchar{.}%
        {\emph{et al}}%
        {\emph{et al.}\@\xspace}%
}
\makeatother

\usepackage{booktabs}
\usepackage{pbox}
\usepackage{multirow}
\usepackage{color}
\usepackage{hhline}
\usepackage{etoolbox}
\usepackage{fixltx2e}

\newcolumntype{R}[1]{>{\raggedright\arraybackslash}p{#1}}
\newcolumntype{L}[1]{>{\raggedleft\arraybackslash}p{#1}}
\newcolumntype{C}[1]{>{\centering\arraybackslash}p{#1}}

\newcommand{\tabincell}[2]{\begin{tabular}{@{}#1@{}}#2\end{tabular}}

\newtoggle{notes}
\togglefalse{notes}

\begin{document}
%
\title{Automatic Face Image Quality Prediction}
%
%
%

\author{Lacey~Best-Rowden,~\IEEEmembership{Student Member,~IEEE,}
        and~Anil~K.~Jain,~\IEEEmembership{Life~Fellow,~IEEE}
\thanks{L. Best-Rowden and A. K. Jain are with the Department
of Computer Science and Engineering, Michigan State University, East Lansing,
MI, 48824.\protect\\
E-mail: \{bestrow1, jain\}@msu.edu}
}

%
%

\markboth{}%
{Shell \MakeLowercase{\textit{et al.}}: Bare Demo of IEEEtran.cls for IEEE Journals}
%



\maketitle

\begin{abstract}
Face image quality can be defined as a measure of the utility of a face image to automatic face recognition. In this work, we propose (and compare) two methods for automatic face image quality based on target face quality values from (i) human assessments of face image quality (matcher-independent), and (ii) quality values computed from similarity scores (matcher-dependent). A support vector regression model trained on face features extracted using a deep convolutional neural network (ConvNet) is used to predict the quality of a face image. The proposed methods are evaluated on two unconstrained face image databases, LFW and IJB-A, which both contain facial variations with multiple quality factors. Evaluation of the proposed automatic face image quality measures shows we are able to reduce the FNMR at 1\% FMR by at least 13\% for two face matchers (a COTS matcher and a ConvNet matcher) by using the proposed face quality to select subsets of face images and video frames for matching templates (i.e., multiple faces per subject) in the IJB-A protocol. To our knowledge, this is the first work to utilize human assessments of face image quality in designing a predictor of unconstrained face quality that is shown to be effective in cross-database evaluation.
\end{abstract}


%
\IEEEpeerreviewmaketitle

\section{Introduction}
%
%
%
%

\IEEEPARstart{T}{he} performance of automatic face recognition systems largely depends on the quality of the face images acquired for comparison. Under controlled image acquisition conditions (\eg ID card face images) with uniform lighting, frontal pose, neutral expression, and standard image resolution, face recognition systems can achieve extremely high accuracies. 
For example, the NIST MBE \cite{nistMBE} reported face verification accuracy of $>$99\% True Accept Rate (TAR) at 0.1\% False Accept Rate (FAR) for a database of visa application photos, and the NIST FRVT 2013 \cite{NIST8009} reported 96\% rank-1 identification accuracy for a database of law enforcement face images (\eg mugshots). 
However, there are many emerging applications of face recognition which seek to operate on face images captured in less than ideal conditions (\eg surveillance) where large intra-subject facial variations are more prevalent, or even the norm, and can significantly degrade the accuracy of face recognition. 
The NIST FRVT 2013 \cite{NIST8009} also demonstrated that mugshot-to-mugshot recognition error rates more than doubled for the top six commercial algorithms when comparing a mugshot gallery to lower quality webcam face images \cite{NIST8009}.

The performance of biometric recognition, in general, is driven by the quality of biometric samples (\eg fingerprint, iris, and face). Biometric sample quality is defined as \emph{a measure of a sample's utility to automatic matching} \cite{GrotherPAMI2007, BharadwajReview,QualityMeasures2012}. A biometric quality measurement should be an indicator of recognition performance where correlation with error rates, such as false non-match rate (FNMR), false match rate (FMR), or identification miss rate, is a desirable property. Essentially, poor quality biometric samples cause a recognition system to fail.

\begin{figure}
\centering
\includegraphics[width=0.98\linewidth]{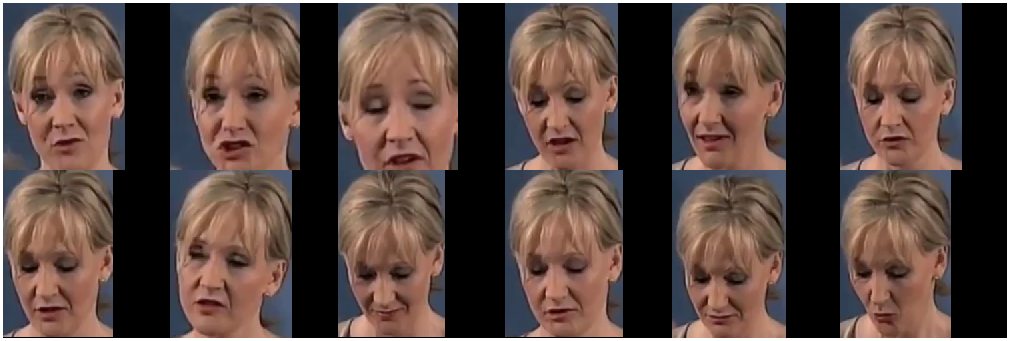}
\caption{Cropped faces from frames of a sample video in the IJB-A \cite{KlareIJBA} unconstrained face database; faces are sorted from high to low face quality by the proposed MQV predictor.}
\label{fig:exVid}
\end{figure}

Adhering to this definition, automatic prediction of \emph{face image quality} (prior to matching and recognition) can be useful for several practical applications. A system with the ability to detect poor quality face images can subsequently process them accordingly. In negative identifications systems (\eg automated security checkpoints at airports), 
persons may intentionally present low quality face images to the system to evade recognition; face quality assessment could flag such attempts and deny services (\eg entry through the checkpoint) until a face image of sufficient quality has been presented.
Face image quality can also be used for quality-based fusion when multiple face images (\eg sequence of video frames, see Fig.~\ref{fig:exVid}) and/or biometric modalities  \cite{PohKittlerPAMI12} (\eg face and fingerprint) of the same subject are available, as well as for 3D face modeling from a collection of face images \cite{Roth3DFace2016}.
Additionally, dynamic recognition approaches \cite{KlareDemographic2012} can make use of face image quality where high-quality face images can be assigned to high-throughput algorithms, while low-quality face images are given to slower, but more robust, algorithms.  


Because biometric sample quality is defined in terms of automatic recognition performance, human visual perception of image quality may not be well correlated with recognition performance \cite{GrotherPAMI2007, BharadwajReview}. 
Particularly, given a fingerprint or iris image, it is difficult for a human to assess the quality in the context of recognition because humans (excluding forensic experts) do not naturally use fingerprints or iris textures for person recognition. 
However, the human visual system is extremely advanced when it comes to recognizing the faces of individuals, a routine daily task. 
In fact, it was recently shown that humans surpass the performance of current state-of-the-art automated systems on recognition of very challenging, low-quality, face images \cite{Blanton2016}.
Even so, to the best of our knowledge, very few studies have investigated face image quality assessment by humans. 
Adler and Dembinsky \cite{AdlerHumanQuality} found very low correlation between human and algorithm measurements of face image quality (98 mugshots of 29 subjects, eight human evaluators), while Hsu \etal \cite{Hsu2006} found some consistency between human perception and recognition-based measures of face image quality (frontal and controlled illumination face images, two human evaluators).

The primary goal of face recognition research is to develop systems which are robust to factors such as pose, illumination, expression, occlusion, resolution, and other intrinsic or extrinsic properties of face images.
Recent works on automatic face recognition have devoted efforts towards recognition of \emph{unconstrained} facial imagery \cite{WangOtto, KlareIJBA, ChenCVPR13, YiScratch_CASIA, ChenDCNN2016, TranGAN2017} where facial variations of any kind can be simultaneously present (\eg face images from surveillance cameras).
However, prior work in face image quality has primarily focused on the quality of lab-collected face image databases (\eg FRGC \cite{FRGC}, GBU \cite{GBU}, Multi-PIE \cite{Multi-PIE}) where facial variations such as illumination and pose are synthetic/staged/simulated in order to isolate and facilitate evaluation of the different quality factors.  

\begin{figure}[t]
\centering
\subfloat[]{
\includegraphics[width=0.95\linewidth]{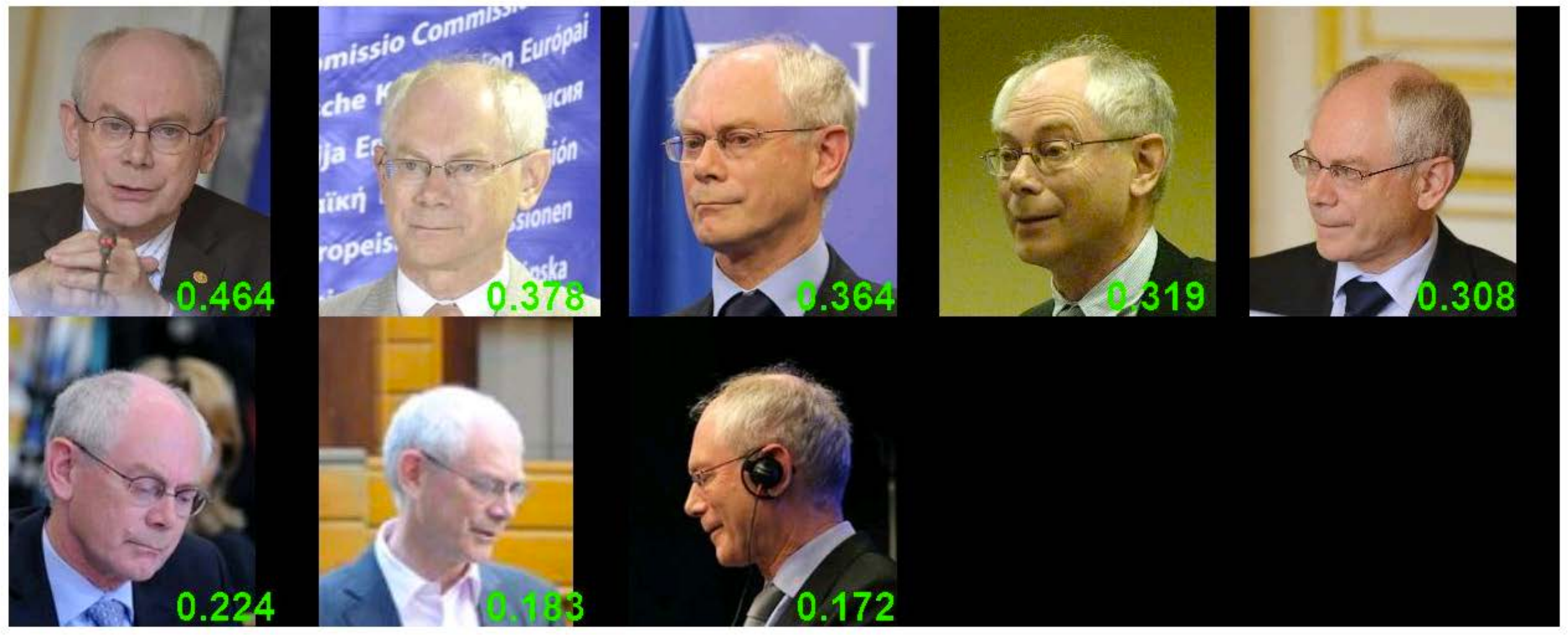}
}\hfill
\subfloat[]{
\includegraphics[width=0.95\linewidth]{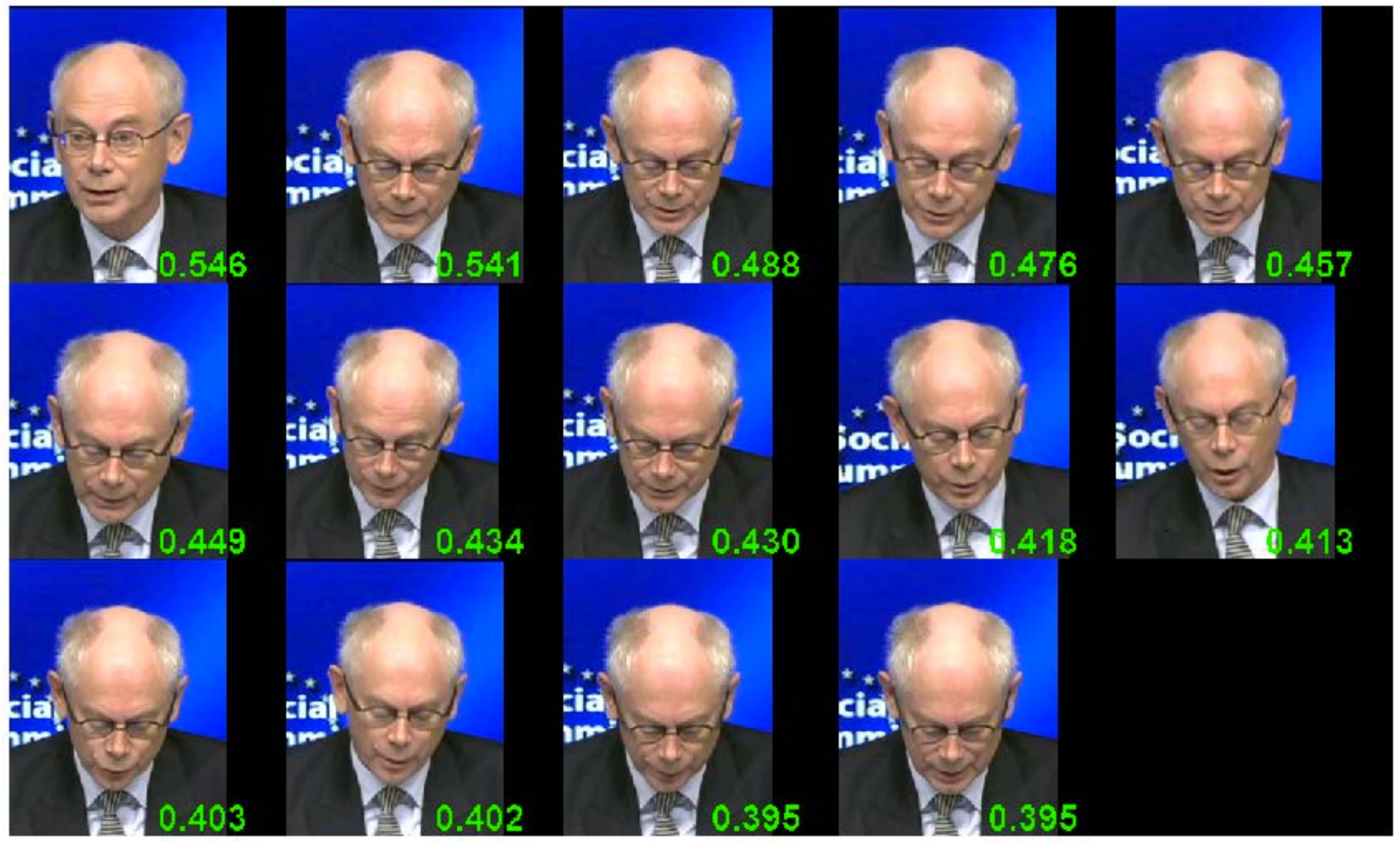}
}\hfill
\subfloat[]{
\includegraphics[width=0.95\linewidth,trim={5mm 0 2mm 5mm},clip]{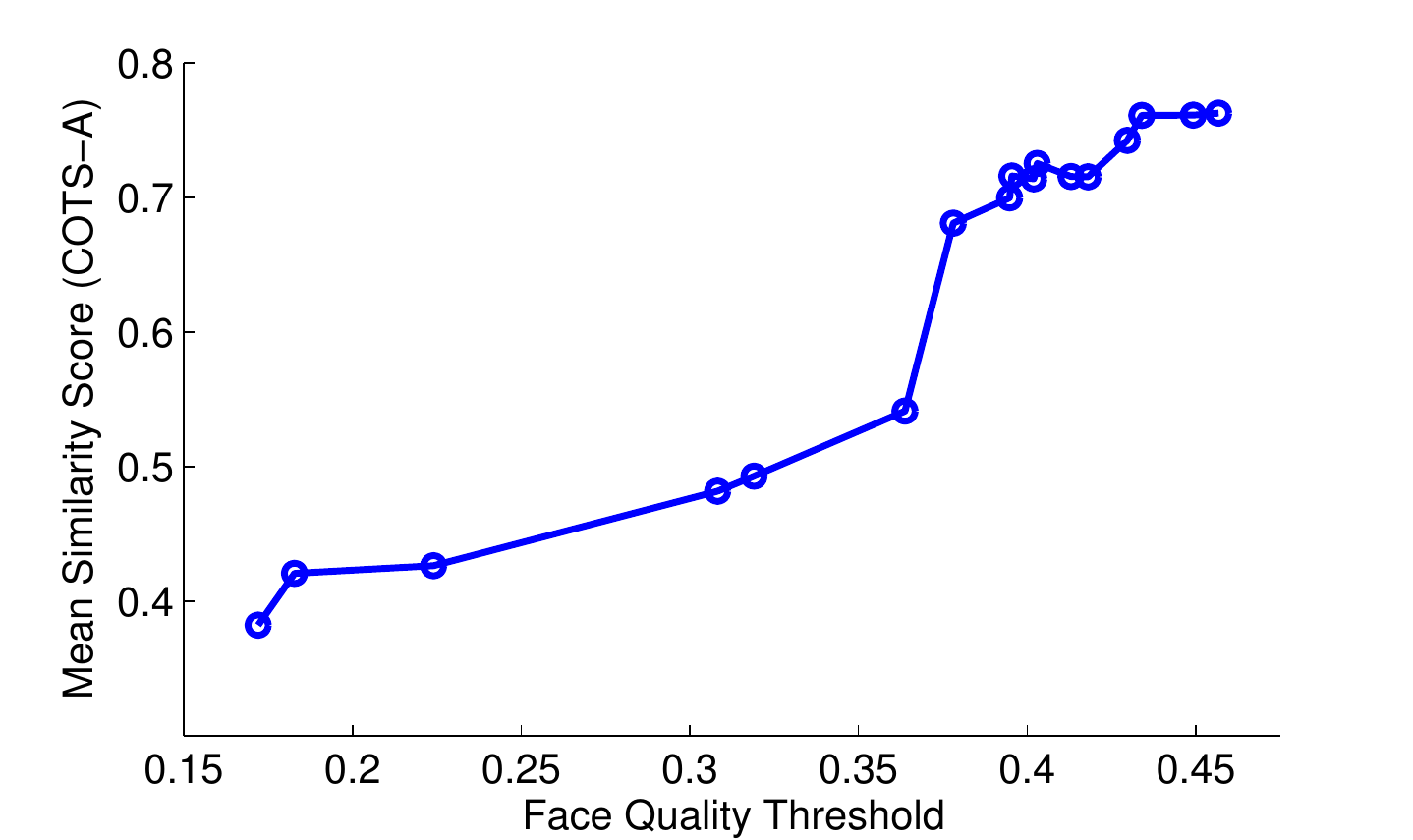}
}
\caption{(a) A gallery and (b) a probe template of the same subject from the IJB-A database \cite{KlareIJBA}. Face image quality values automatically predicted by the proposed HQV predictor are given in green (lower value indicates lower face quality). To obtain a single similarity score for the multiple faces in the gallery and probe templates, score-level fusion is typically the baseline approach. (c) Score-level fusion (mean rule) of COTS-A similarity scores using only those faces with face quality above a threshold increases the fused similarity score as the threshold increases. In this scenario, a monotonically increasing relationship is desired between the mean genuine similarity score and the face quality threshold because higher quality faces should result in higher genuine similarity.}
\label{fig:exTemplateMatching}
\end{figure}

In this work, we focus on automatic face image quality of unconstrained face images using the Labeled Faces in the Wild (LFW) \cite{LFWTech} and IARPA Janus Benchmark A (IJB-A) \cite{KlareIJBA} unconstrained face datasets.
The contributions of this work are summarized as follows:
\begin{itemize}
\item Collection of human ratings of face image quality for a large database of unconstrained face images (namely, LFW \cite{LFWTech}) by crowdsourcing a small set of pairwise comparisons of face images and inferring the complete ratings with matrix completion \cite{Yi2013}. 
\item Investigation of the utility of face image quality assessment by humans in the context of automatic face recognition performance. This is the first study on human quality assessment of face images that exhibit a wide range of quality factors (\ie \emph{unconstrained} face images). 
\item Comparison of two methods for establishing the target quality values of face images in a database based on: (i) human quality ratings (\emph{matcher-independent}) and (ii) quality values computed from a similarity scores obtained from face matchers (\emph{matcher-dependent}). The latter serves as an ``oracle'' for a face quality measure that is correlated with recognition performance. 
\item A model for automatic prediction of face image quality of independent test images (not seen during training) using face features from a learned deep neural network extracted prior to matching. 
\end{itemize}
Our experimental evaluation follows the methodology advocated by Grother and Tabassi \cite{GrotherPAMI2007} where a biometric quality measurement  is tested by ``relating quality values to empirical matching results.''
The quantitative evaluation presented is aimed at the application of using the face image quality measures to improve error rates (\eg FNMR) of automatic face recognition systems by rejecting low-quality face images. For example, in template-based matching (\eg the IJB-A protocol \cite{KlareIJBA}) standard score-level fusion over multiple faces per subject can be improved by removing low-quality faces prior to computing the mean of the similarity scores (see Fig.~\ref{fig:exTemplateMatching}).

\begin{table*}[t]
\centering
\footnotesize
\renewcommand{\arraystretch}{1.3}
\caption[caption]{Summary of Related Work on Automatic Methods for Face Image Quality}
\label{tab:relatedWorkSmall}
\begin{tabular}{|R{2cm}|R{3cm}|R{2.75cm}|R{4.3cm}|R{3.5cm}|}
\hline
\multicolumn{1}{|c|}{\tabincell{c}{\textbf{Study}\\\textbf{(year)}}} & \multicolumn{1}{c|}{\tabincell{c}{\textbf{Database:}\\\textbf{Num. of images (subjects)}}} & \multicolumn{1}{c|}{\tabincell{c}{\textbf{Target Quality}\\\textbf{Value}}} & \multicolumn{1}{c|}{\textbf{Learning Approach}} & \multicolumn{1}{c|}{\textbf{Evaluation}}\\
\hline\hline
Hsu \etal \cite{Hsu2006} (2006) & 
\raisebox{-4mm}{\tabincell{l}{FRGC: 1,886 (n/a)\\passports: 2,000 (n/a)\\mugshots: 1,996 (n/a)}} & 
Continuous (genuine score) & Neural network to combine 27 quality measures (exposure, focus, pose, illumination, etc.) for prediction of genuine scores & ROC curves for different levels of quality (FaceIt algorithm by Identix) \\ \hline
Aggarwal \etal \cite{AggarwalMDS2011} (2011) & 
\raisebox{-2mm}{\tabincell{l}{Multi-PIE: 6,740 (337)\textsuperscript{$\ast$}\\FacePix: 1,830 (30)}} & 
Continuous (genuine score) or Binary (algorithm success vs. failure; requires matching prior to quality) & MDS to learn a mapping from illumination features to genuine scores. Predicted genuine score compared to algorithm score to decide algorithm success or failure & Prediction accuracy of algorithm success vs. failure, ROC curves for predicted, actual, 95\% and 99\% retained (SIFT-based and PittPatt algorithms)\\ \hline
Phillips \etal \cite{PhillipsBTAS2013} (2013) &
 \raisebox{-2mm}{\tabincell{l}{PaSC: 4,688 (n/a)\\GU\textsuperscript{\dag}: 4,340 (437)}} & 
 Binary (low vs. high) & PCA + LDA classifier  & Error vs. Reject curve for FNMR vs. percent of images removed\\ \hline
Bharadwaj \etal \cite{BharadwajICIP2013} (2013) & 
\raisebox{-2mm}{\tabincell{l}{CAS-PEAL: n/a (1,040)\\SCface: n/a (130)}} & 
Quality bins (poor, fair, good, excellent) & Multi-class SVM trained to predict face quality bin from holistic face features (GIST and HOG) & ROC curves, rank-1 accuracy, EER, \% histogram overlap (COTS algorithm)\\ \hline
Abaza \etal \cite{AbazaIET2014} (2014) & 
\raisebox{0mm}{\tabincell{l}{GU\textsuperscript{\dag}: 4,340 (437) }} & 
Binary (good vs. ugly) & Neural network (1-layer) to combine contrast, brightness, sharpness, focus, and illumination measures & Rank-1 identification for blind vs. quality-selective fusion\\ \hline
Dutta \etal \cite{DuttaBayesianIJCB2014} (2014) & 
\raisebox{0mm}{\tabincell{l}{Multi-PIE: 3,370 (337)\textsuperscript{\ddag}}} & 
Continuous (false reject rate) & Probability density functions (PDFs) model interaction between image quality (deviations from frontal and uniform lighting) and recognition performance & Predicted vs. actual verification performance for different clusters of quality (FaceVACS algorithm)\\ \hline
Kim \etal \cite{Kim2015} (2015) & 
\raisebox{0mm}{\tabincell{l}{FRGC: 10,448 (322)}} & 
Binary (low vs. high) or Continuous (confidence of the binary classifier) & Objective (pose, blurriness, brightness) and Relative (color mismatch between train and test images) face image quality measures as features fed into AdaBoost binary classifier & Identification rate w.r.t. fraction of images removed, ROC curve with and without low quality images (SRC face recognition algorithm)\\ \hline
Chen \etal \cite{Chen2015} (2015) & 
\raisebox{-6mm}{\tabincell{l}{SCface: 2,080 (130)\\(trained with FERET,\\FRGC, LFW, and\\non-face images)}} & 
0 -- 100 (rank-based quality score) & A ranking function is learned by assuming images from different databases are of different quality and images from same database are of equal quality & Visual quality-based rankings, Identification rate (Gabor filter based matcher)\\ \hline
Proposed Approach & \raisebox{-6mm}{\tabincell{l}{LFW: 13,233 (5,749) for\\training and testing\\ IJB-A: 5,712 images and\\ 2,085 videos (500) for testing}} & Continuous (human quality ratings or normalized comparison scores) & Support vector regression with image features from a deep convolutional neural network \cite{WangOtto} & Error vs. Reject curves, visual quality-based ranking\\ \hline
\multicolumn{5}{l}{\emph{Note:} n/a \emph{indicates that the authors did not report the number of images or subjects (an unknown subset of the database may have been used).}}\\
\multicolumn{5}{l}{\textsuperscript{$\ast$}Only the illumination subset of Multi-PIE database \cite{Multi-PIE} was used.}\\
\multicolumn{5}{l}{\textsuperscript{\dag}GU denotes the Good and Ugly partitions of the Good, Bad, and Ugly (GBU) face database \cite{GBU}.}\\
\multicolumn{5}{l}{\textsuperscript{\ddag}Only neutral expression face images from Multi-PIE database \cite{Multi-PIE} were used.}\\
\end{tabular}
\end{table*}

\section{Related Work}
\label{sec:related}

A number of studies (\eg \cite{BeveridgeFRVT, BeveridgeGLMMFRGC09, AbazaIET2014}) have offered in depth analyses of the performance of automatic face recognition systems with respect to different \emph{covariates}. 
These studies have identified key areas of research and have guided the community to develop algorithms that are more robust to the multitude of variations in face images. 
The covariates studied include \emph{image-based}, such as pose, illumination, expression, resolution, and focus, as well as \emph{subject-based}, such as gender, race, age, and facial accessories (\eg eyeglasses). 
In general, it is typically shown that face recognition performance degrades due to these different sources of variability. Intuitively, the magnitude of degradation is algorithm-specific. 

Prior works have proposed face image quality as some measure of the similarity to reference, or ``ideal", face images (typically frontal pose, uniform illumination, neutral expression). For example, \cite{SellahewaAdaptiveFusion2010} uses luminance distortion from a high quality reference image for adaptive fusion of two face representations. Wong \etal \cite{WongPatchBased2011} propose probabilistic similarity to a reference model of ``ideal" face images for selecting high quality frames in video-to-video verification, and Best-Rowden \etal \cite{BestRowdenTIFS} investigated structural similarity (SSIM) for quality-based fusion within a collection of face media.
Reference-based approaches are dependent on the face images used as reference and may not generalize well to different databases or face images with multiple quality factors present. 


More recently, especially with the influx of unconstrained face images, interest has peaked in automatic measures for face image quality that can encompass multiple quality factors, and hence, determine the degree of suitability for automatic matching of an arbitrary face image. Table~\ref{tab:relatedWorkSmall} summarizes related works in automatic face image quality which are learning-based approaches. These methods are related in that they all define some target quality which is related to automatic recognition performance. The target quality value can be a prediction of the genuine score \cite{Hsu2006, AggarwalMDS2011}, a bin indicating that an image is poor, fair, or good for matching \cite{BharadwajICIP2013}, or a binary value of low vs. high quality image \cite{PhillipsBTAS2013, AbazaIET2014, Kim2015}. 
For example, Bharadwaj \etal fuse similarity scores from two COTS matchers, define quality bins based on CDFs of images that were matched correctly and incorrectly, and use a support vector machine (SVM) trained on holistic image features to classify the quality bin of a test image \cite{BharadwajICIP2013}. Rather than defining target quality values for a training database of face images, Chen \etal propose a learning to rank framework which assumes (i) a rank-ordering ($\prec$) of a set of databases, such that (non-face images) $\prec$ (unconstrained face images) $\prec$ (ID card face images), and (ii) face images from the same database have equal quality; rank weights from five different image features are learned and then mapped to a quality score 0$-$100 \cite{Chen2015}.

In our approach, we establish the target quality values (defined as either human quality ratings or score-based values from a face matcher) of a large database of unconstrained face images, extract image features using a deep ConvNet \cite{WangOtto}, and learn a model for prediction of face quality from the ConvNet features using support vector regression (SVR). The target quality values in this work are continuous and allow for a fine-tuned quality-based ranking of a collection of face images.

\iftoggle{notes}{
\textcolor{red}{[Biometric Completeness, When High Quality Images Match Poorly, etc.}
\textcolor{red}{The emerging body of literature on post-recognition score analysis has been largely constrained to biometrics, where the analysis has been shown to successfully complement or replace image quality metrics as a predictor. [Meta-Recognition: The Theory and Practice of Recognition Score Analysis, W. Schemer \etal, PAMI 2011]}
}

\section{Face Databases and COTS Matchers}
\begin{figure}
\centering
\subfloat[]{
\hspace{-2mm}
\includegraphics[width=0.48\linewidth,trim={0 0 8.9cm 0},clip]{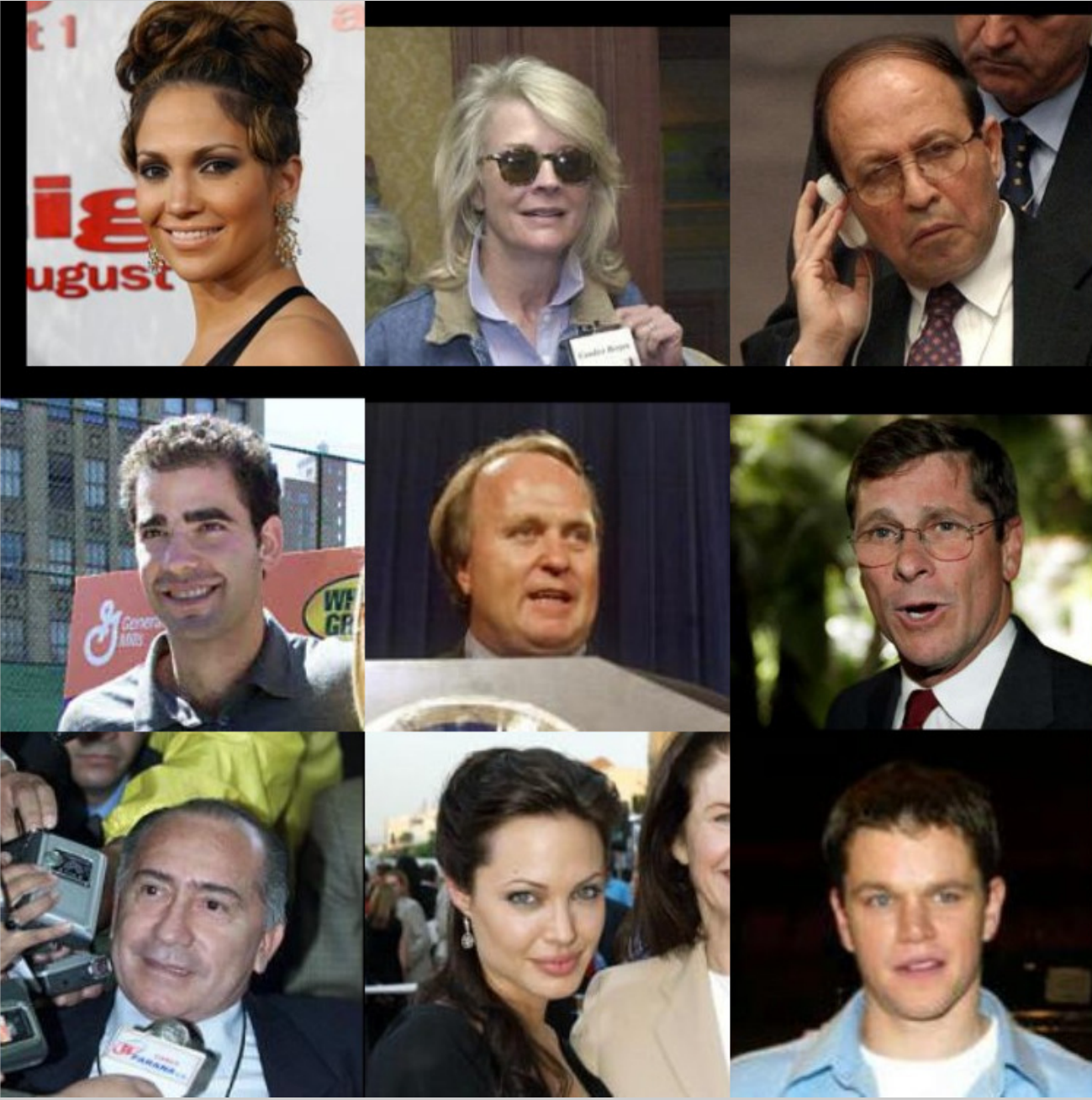}\hfill
}
\subfloat[]{
\includegraphics[width=0.48\linewidth,trim={8.9cm 0 0 0},clip]{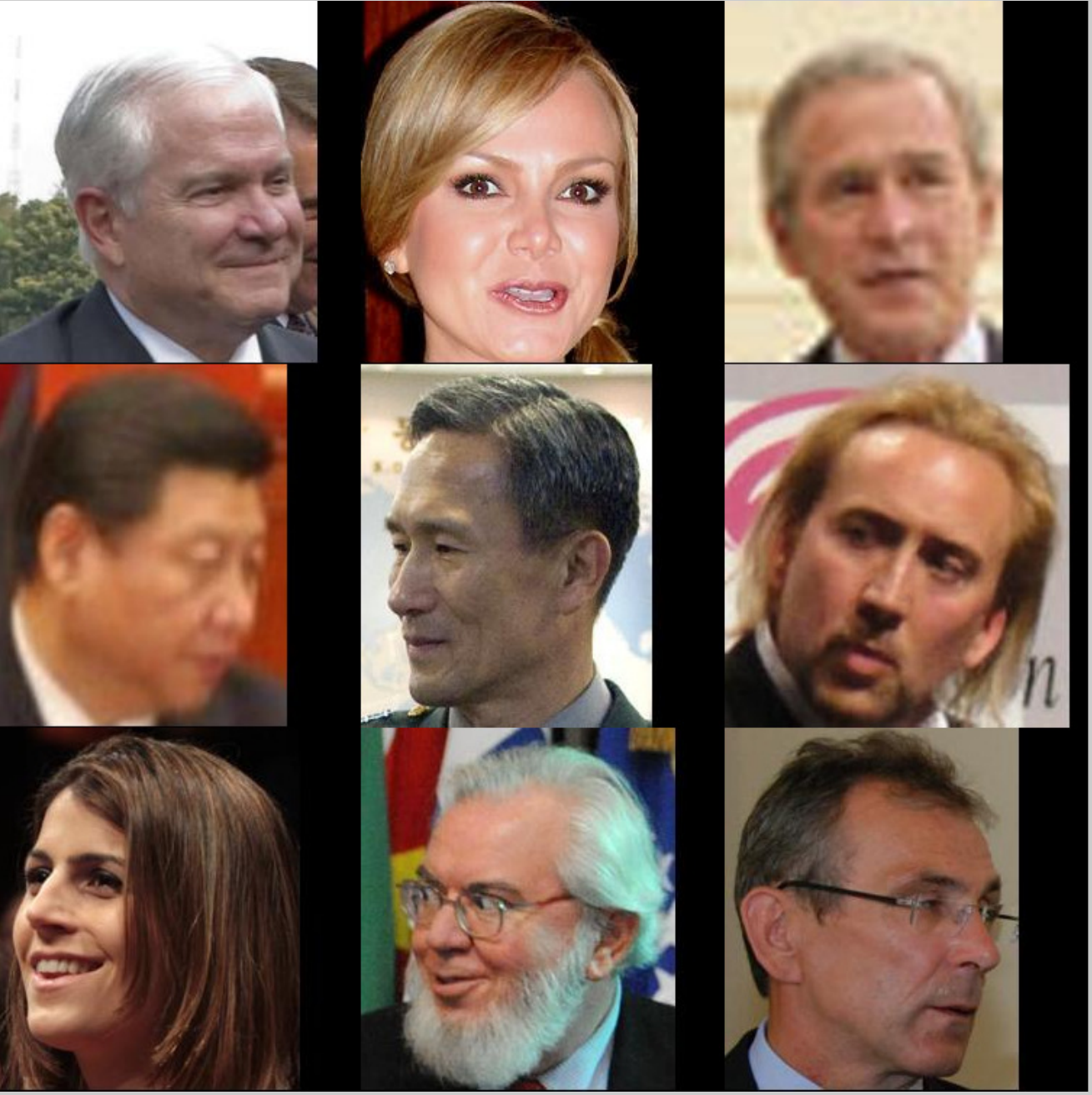}
}
\caption{Sample face images from the (a) LFW \cite{LFWTech} and (b) IJB-A \cite{KlareIJBA} unconstrained face databases.}
\label{fig:lfw_ijba}
\end{figure}

We utilize two unconstrained face databases: Labeled Faces in the Wild (LFW) \cite{LFWTech} and IARPA Janus Benchmark A (IJB-A) \cite{KlareIJBA}. Both LFW and IJB-A contain face images with unconstrained facial variations that affect the performance of face recognition systems (\eg pose, expression, illumination, occlusion, resolution, etc.). The LFW database consists of 13,233 images of 5,749 subjects, while the IJB-A database consists of 5,712 images and 2,085 videos of 500 subjects. 
Face images in the LFW database were detected by the Viola-Jones face detector \cite{ViolaJones} so the pose variations are limited by the pose tolerance of the Viola-Jones detector. Face images in IJB-A were manually located, so the database is considered more challenging than LFW due to full pose variations \cite{KlareIJBA}. See Fig.~\ref{fig:lfw_ijba} for sample face images from the two databases.

Because face image quality needs to be evaluated in the context of automatic face recognition performance, we make use of two commercial face matchers, denoted as COTS-A\footnote{COTS-A was one of the top performers in the NIST Face Recognition Vendor Test (FRVT) \cite{NIST8009}.} and COTS-B. Table~\ref{tab:blufr} shows that COTS-A and COTS-B are competitive algorithms on the BLUFR protocol \cite{LiaoBenchmark} for the LFW database. Performance is also reported for the deep learning-based matcher, denoted ConvNet, proposed by Wang \etal \cite{WangOtto}. The 320-dimensional feature representation output by a single deep convolutional neural network from \cite{WangOtto} is used in this work to predict face image quality. 

\iftoggle{notes}{
\subsection{\textcolor{blue}{ConvNet:}}
Network architecture and training:
10 convolutional layers with filters of size 3 $\times$ 3, outputs a 320-dimensional feature vector. 

Face detection and alignment:
DLIB implementation of ensemble of regression trees method, 68 facial landmarks detected, face rotated upright based on eye locations, central point of the face is mid-point between leftmost and rightmost landmarks used to horizontally center the face, the center point between the two eyes is placed 45\% from the top and the center point of the mouth is placed 25\% from the bottom of the image.
Images are resized to 110 $\times$ 110 pixels.
``well-aligned'' vs. ``poorly-aligned'' images

Set-to-set matching for IJB-A template protocol:
If well-aligned images are available, poorly-aligned images are discarded. 
Average score over all well-aligned images (if available); otherwise, average score over all poorly-aligned images. 
}

\begin{table}
\renewcommand{\arraystretch}{1.1}
\caption{Verification and Open-Set Identification Performance of Various Face Recognition Algorithms on the LFW Database \cite{LFWTech} Under the BLUFR Protocol \cite{LiaoBenchmark} }
\centering
\small
\begin{tabular}{| m{3.7cm}  |C{1.4cm} | C{1.4cm} |}
\hline
\multicolumn{1}{|C{3.7cm}|}{Algorithm}	& \multicolumn{1}{C{1.4cm}|}{TAR @ 0.1\% FAR} & \multicolumn{1}{C{1.3cm}|}{DIR\textsuperscript{$\dag$} @ 1\% FAR} \\
\hline\hline
HDLBP $+$ JointBayes \cite{ChenCVPR13}\textsuperscript{$\ast$} & 41.66 & 18.07\\
Yi \etal \cite{YiScratch_CASIA} & 80.26 & 28.90\\
\hline
ConvNet \cite{WangOtto} (\# nets = 1) & 85.00 & 49.10\\
ConvNet \cite{WangOtto} (\# nets = 9) & 89.80 & 55.90 \\
COTS-A & 88.14 & 76.28\\ 
COTS-B & 76.01 & 53.21\\ 
\hline
\multicolumn{3}{R{6.5cm}}{\scriptsize \textsuperscript{$\ast$}Performance here for \cite{ChenCVPR13} was reported by \cite{LiaoBenchmark}}\\
\multicolumn{3}{R{6.5cm}}{\scriptsize \textsuperscript{$\dag$}DIR = Detection and Identification Rate}\\
\end{tabular}
\label{tab:blufr}
\end{table}


\section{Face Image Quality Labels}
\label{sec:qualityLabels}
Biometrics and computer vision heavily rely on supervised learning techniques when training sets of \emph{labeled} data are available. 
When the aim is to develop an automatic method for face image quality, compiling a quality-labeled face image database is not straightforward.
The definition of face image quality (\ie a predictor of automatic matching performance) does not lend itself to explicit labels of face image quality, unlike labels of facial identity or face vs. non-face labels for face recognition and detection tasks, respectively.
Possible approaches for establishing target quality labels of face images include:
\begin{enumerate}
\item Combine various measurements of image quality factors into a single value which indicates the overall face quality.
\item Human annotations of perceived image quality. 
\item Use comparison scores (or performance measures) from automatic face recognition matchers. 
\end{enumerate}

The issue with 1) is that it is an \emph{ad-hoc} approach and, thus far, has not achieved much success (see \cite{PhillipsBTAS2013}). 
The issue with 2) is that human perception of quality may not be indicative of automatic recognition performance; previous works \cite{GrotherPAMI2007, BharadwajICIP2013} have stated this conjecture, but to our knowledge, the only studies to investigate human perception of face quality were conducted on constrained face images (\eg mugshots) \cite{AdlerHumanQuality, Hsu2006}. 
The issue with 3) is that comparison scores are obtained from a \emph{pair} of images, so labeling single images based on comparison scores (or performance) can be problematic. However, this approach achieved some success for fingerprint quality \cite{NFIQ, GrotherPAMI2007}, and only few studies \cite{BharadwajICIP2013, PhillipsBTAS2013} have considered it for face quality. In this work, we investigate both methods 2) and 3), detailed in the remainder of this section.

\subsection{Human Quality Values (HQV)}
Because of the inherent ambiguity in the definition of face image quality, framing an appropriate prompt to request a human to label the quality of a face image is challenging. If asked to rate a face image on a scale of 1 to 5, for example, there are no notions as to the meaning of the different quality levels. Additionally, some prior exposure to the variability in the face images that the human will encounter may be necessary so that they know what kinds of ``quality" to expect in face images (\ie a baseline) before beginning the quality rating task. 

In this work, we choose to only collect quality labels for relative pairwise comparisons of face images by asking the following question: ``Which face (left or right) has better quality?" 
Crowdsourcing literature \cite{Yi2013} has demonstrated that ordinal (comparison-based) tasks are generally easier for participants and take less time than cardinal (score-based) tasks. Ordinal tasks additionally avoid calibration efforts needed for cardinal responses from raters inherently using different ranges for decision making (\ie biased ratings, inflated vs. conservative ratings, changes in absolute ratings over time with exposure to more data). 

Given the collected pairwise face comparisons, to obtain absolute quality ratings for \emph{individual} face images, we make use of a matrix completion approach \cite{Yi2013} to infer the quality rating matrix from the pairwise comparisons. Because it is infeasible to have multiple persons manually assess and label the qualities of \emph{all} face images in a large database, this approach is desirable in that it only requires a small set of pairwise quality labels from each human rater in order to infer the quality ratings for the entire database. The details of data collection and the matrix completion approach are discussed in the remainder of this section.


\subsubsection{Crowdsourcing Comparisons of Face Quality}
\begin{figure}[!t]
\centering
\includegraphics[width=\linewidth,trim={28mm 5mm 28mm 0},clip]{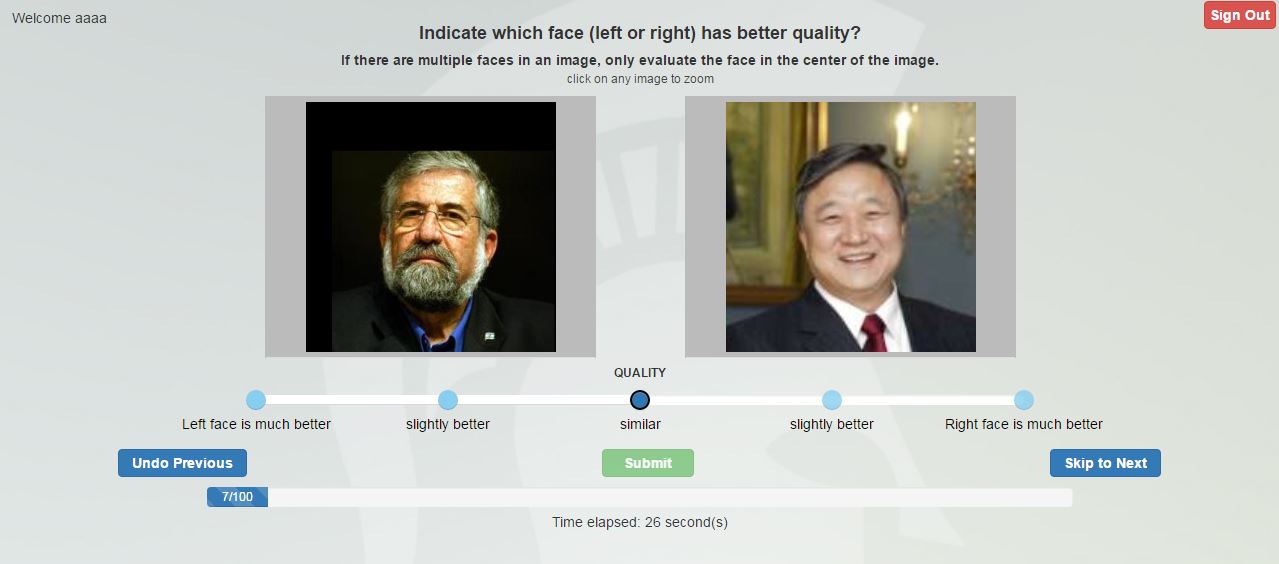}
\caption{The interface used to collect responses for pairwise comparisons of face image quality from MTurk workers.}
\label{fig:FaceOFF}
\end{figure}

Amazon Mechanical Turk (MTurk)\footnote{\url{https://www.mturk.com}} was utilized to facilitate collection of pairwise comparisons of face image quality from multiple human raters (\ie MTurk ``workers'').
Given a pair of face images, displayed side by side, our Human Intelligence Task (HIT) was to respond to the prompt ``Indicate which face has better quality'' by selecting one of the following: 
(i) left face is much better, (ii) left face is slightly better, (iii) both faces are similar, (iv) right face is slightly better, and (v) right face is much better.
Fig.~\ref{fig:FaceOFF} shows the interface used to collect the responses.\footnote{The tool is available at \url{http://cse.msu.edu/~bestrow1/FaceOFF/}.} 

\begin{figure}[!t]
\centering
\includegraphics[width=0.975\linewidth]{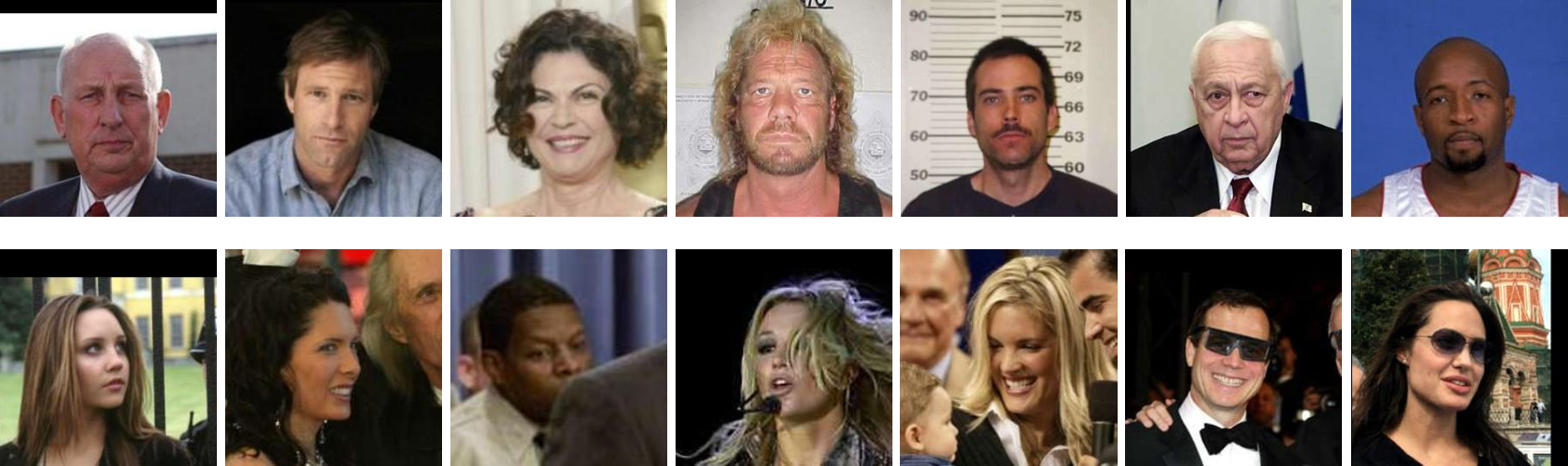}
\caption{Face images (from the LFW database) selected for the 6 tutorial pairs which are used to check whether MTurk workers understood the task before completing the pairwise comparisons used in our study of face image quality. For each of the 6 tutorial pairs, one image was selected from the top row (high quality images) and one image was selected from the bottom row (low quality images), so the pairwise comparison of face quality had an unambiguous answer. }
\label{fig:tutorialImages}
\end{figure}

Our HIT requested each worker to provide responses to a total of 1,001 face image pairs, made up of 6 tutorial pairs, 974 random pairs, and 21 consistency check pairs.
The tutorial pairs were pre-selected from the LFW database where the quality of one image was clearly better than the quality of the other (Fig.~\ref{fig:tutorialImages} shows the sets of images used). Because we expected these easy pairs to elicit ``correct" responses, they allowed us to ensure that the worker had completed the tutorial introduction and understood the goal of the task. The next 974 pairs of images were chosen randomly from the LFW database, while the final 21 pairs were selected from the set of 974 as repeats to test the consistency of the worker's responses. 
MTurk workers who attempted our HIT were only allowed to complete it if they passed the tutorial pairs, and we only accepted the submitted responses from workers who were consistent on at least 10 out of the 21 consistency check pairs.  

In order to be eligible to attempt our HIT for assessment of face image quality, MTurk workers had to have previously completed at least 10,000 HITs from other MTurk ``requesters'' with an approval rating of at least 99\%. These stringent qualifications helped to ensure that only experienced and reliable workers (in terms of MTurk standards) participated in our data collection.\footnote{The MTurk worker qualifications are managed by the MTurk website.} 
A total of 435 MTurk workers began our HIT. After removing 245 workers who did not complete the full set of 1,001 pairwise comparisons and 4 workers who failed the consistency check (inconsistent response for 10 or more of the 21 repeated pairs), a total of 194 workers were each compensated US \$5.00 through the MTurk crowdsourcing service. In total, this quality labeling costed less than US \$1,000 and all HITs were completed in less than one day.


\subsubsection{Matrix Completion}
After collecting random sets of pairwise comparisons of face image quality from 194 workers via MTurk, we use the matrix completion approach proposed by Yi \etal \cite{Yi2013} to infer a complete set of quality ratings for each worker on the entire LFW database (13,233 total face images). 
The aim is to infer ${\hat{F} \in \mathbb{R}^{n\times m}}$, the worker-rating matrix for face image qualities, where $n$ is the number of workers and $m$ is the number of face images. 

Yi \etal \cite{Yi2013} show
\iftoggle{notes}{
\textcolor{blue}{[this result is from a different paper cited in Yi \etal. Perhaps we should say, "Yi et al. [] stated that... and then include the citation of the paper where this result was derived.]}
\textcolor{red}{with a low-rank assumption on F ... exploit pairwise information obtained from a crowd of users.}
}
that only ${O(r \log m)}$ pairwise queries are needed to infer the full ranking list of a worker for all $m$ items (face images), where $r$ is the rank of the unknown rating matrix (${r \ll m}$). The maximum possible rank of the unknown rating matrix is ${r = n = 194}$ (number of workers), so ${O(194 \log 13,233) \approx 800}$; hence, the $974$ random pairs per worker collected in our study are sufficient to do the matrix completion, especially since we expect $r < n$ (\ie the quality ratings from the $n$ workers are not all independent).

\begin{figure}[!t]
\centering
\includegraphics[width=0.9\linewidth]{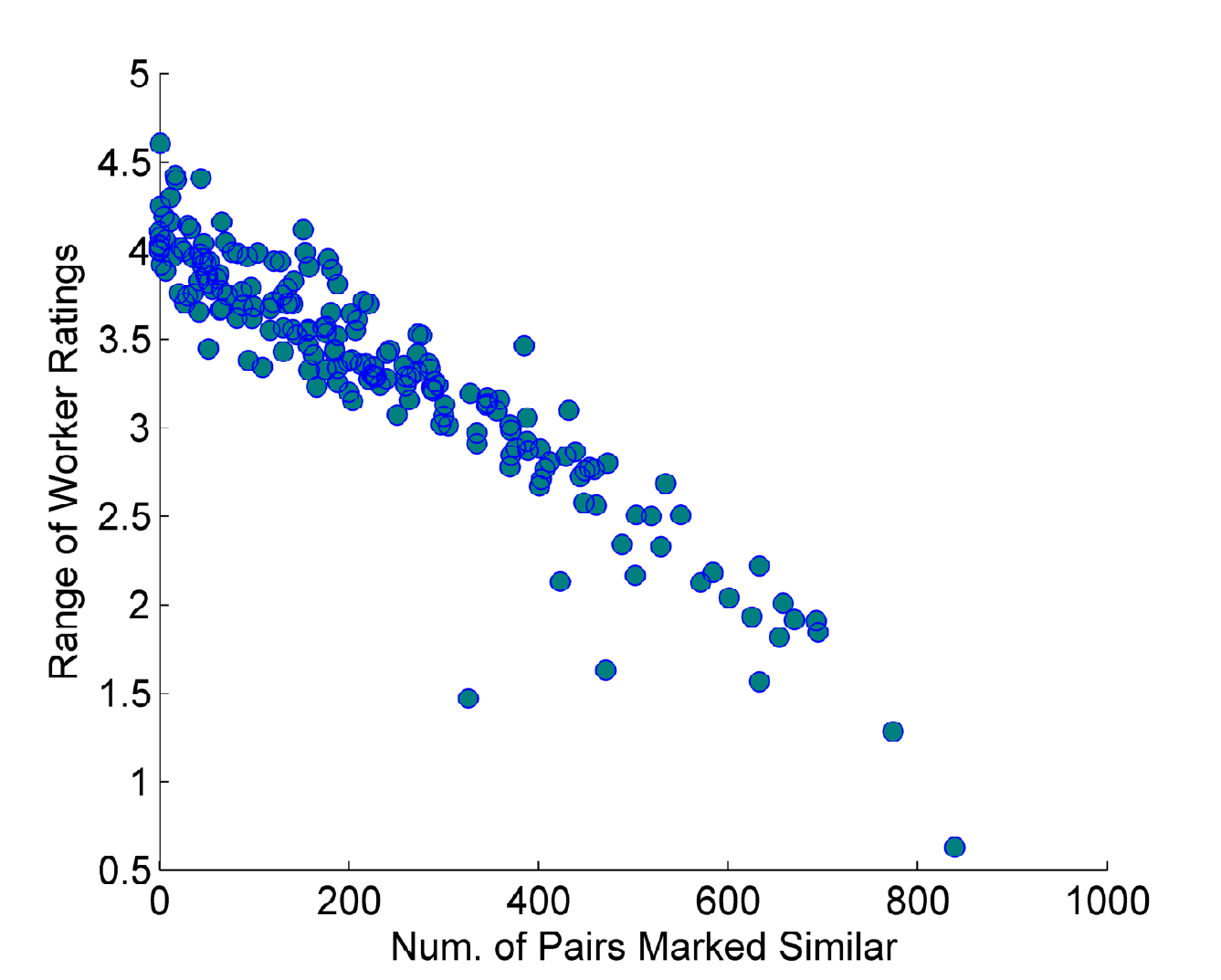}
\caption{The resulting range of the face quality values (after matrix completion) for a particular worker inversely depends on the number of pairs that the worker marked ``Similar'' quality, rather than choosing left/right image is better quality. 
}
\label{fig:ratingsRange}
\end{figure}

While relative pairwise comparisons are often preferred in crowd-based tasks \cite{Yi2013} because they avoid biases from raters' tendencies to give conservative or inflated responses when using an absolute scale (\eg quality levels 1 to 5), we still observed a bias after the matrix completion where the bias is from a tendency to respond ``Similar''. Fig.~\ref{fig:ratingsRange} shows an inverse relationship between the number of pairs that a worker marked ``Similar" and the resulting range of quality ratings for that worker (after matrix completion). Note that this bias is not due to the coarse levels of left (right) image is ``much better'' vs. ``slightly better'' because prior to matrix completion we combine these responses to simply ``left (right) is better''. Because of this observation, \emph{min-max} normalization was performed on each worker's quality ratings to transform to the same range (0 to 1).

\begin{figure}[!t]
\centering
\includegraphics[width=0.9\linewidth]{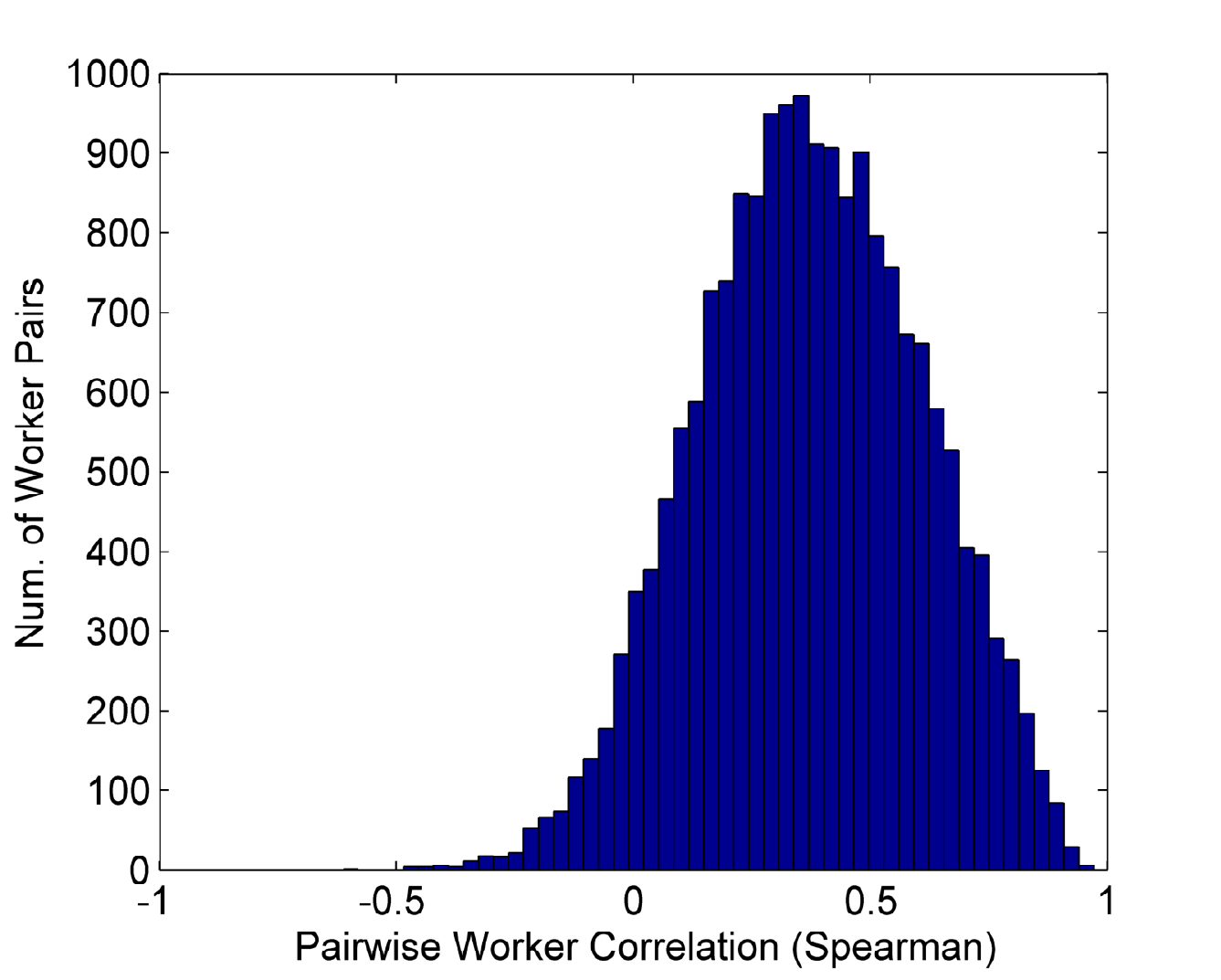}
\caption{Histogram of Spearman rank correlation between the face image quality ratings of all pairs of MTurk workers (${194 \choose 2}=18,721$ total pairs of workers). The face quality ratings are those obtained after matrix completion. The degree of concordance between workers is $0.37$, on average, indicating significant variability in the face quality ratings from different workers.}
\label{fig:workerCorrelation}
\end{figure}

After matrix completion, the worker-rating matrix ${\hat{F}}$ contains face image quality ratings from 194 different workers for each face image in the LFW database.
Figure~\ref{fig:workerCorrelation} shows there is significant variability in the face quality ratings from different workers.  
With the aim of obtaining a single quality rating per face image in the LFW database, we simply take the \emph{median} value from all 194 workers to reduce the ${194\times13,233}$ matrix of quality ratings to a ${1\times13,233}$ vector of quality ratings (one per image in LFW). We empirically tested other heuristics (mean, min, max) but found that median seemed to result in the best quality ratings. 


\subsection{Matcher Quality Values (MQV)}
\label{sec:mqv}
Target quality labels acquired from similarity scores serve as an ``oracle" for a quality measure that is highly correlated with automatic recognition performance. For example, if the goal is to detect and remove low-quality face images to improve the FNMR, then face images should be removed from a database in the order of their genuine comparison scores. 
\iftoggle{notes}{
\textcolor{blue}{[But predicting the quality as the genuine score is equivalent to the recognition problem... ]}
}
Previous works on biometric quality (fingerprint \cite{NFIQ, GrotherPAMI2007} and face \cite{BharadwajICIP2013}) have defined ``ground truth'' or ``target'' quality labels as a measure of the separation between the sample's genuine score and its impostor distribution when compared to a gallery of enrollment samples. A normalized comparison score for the $j$th query sample of subject $i$ can be defined as,
\begin{equation}
\label{eqn:zij}
z_{ij} = (s^G_{ij} - \mu^I_{ij})/\sigma^I_{ij},
\end{equation}
where $s^G_{ij}$ is the genuine score and $\mu^I_{ij}$ and $\sigma^I_{ij}$ are the mean and standard deviation, respectively, of the impostor scores for the query compared to the gallery. 
Previous works then bin the $Z$-normalized comparison scores into quality bins based on the cumulative distribution functions (CDFs) of sets of correctly and incorrectly matched samples \cite{NFIQ, GrotherPAMI2007, BharadwajICIP2013}. Instead, we propose to directly predict $z_{ij}$ for a given face image to obtain a continuous measure of face image quality. 

Target quality values defined based on comparison scores are confounded by the fact that a comparison score is computed from \emph{two} face images, but the aim is to label the quality of a \emph{single} face image. Because comparison scores are typically governed by low-quality samples \cite{GrotherPAMI2007}, the quality value can be assigned to the probe image under the simplifying assumption that the quality of an enrollment image is at least as good as the quality of a probe image.

To allow for this simplifying assumption, we manually selected the best quality image available for the 1,680 subjects in the LFW database with at least two face images. The best image selected by us is placed in the gallery (1,680 images, one per subject), while the remaining 7,484 images of these subjects are used as the probe set. The additional 4,069 images in the LFW database (subjects with only a single image) are used to extend the size of the gallery. 
Normalized comparison scores, $z_{ij}$, are computed using Eqn. (\ref{eqn:zij}) for the 7,484 probe images for each of the face matchers (COTS-A, COTS-B, and ConvNet) and used as score-based target face quality values for learning the face quality predictor. 

\iftoggle{notes}{
\textcolor{blue}{\textbf{Pairwise Quality:} Beveridge \etal \cite{BeveridgeFG2011} argue that face image quality is not an intrinsic property; if this were true, high-quality face images would consistently match correctly and low-quality face images would consistently match incorrectly.
Instead, they observe that verification can be correct if both images lie in the same quality space.}
}

\begin{figure}[!t]
\centering
\includegraphics[width=0.8\linewidth]{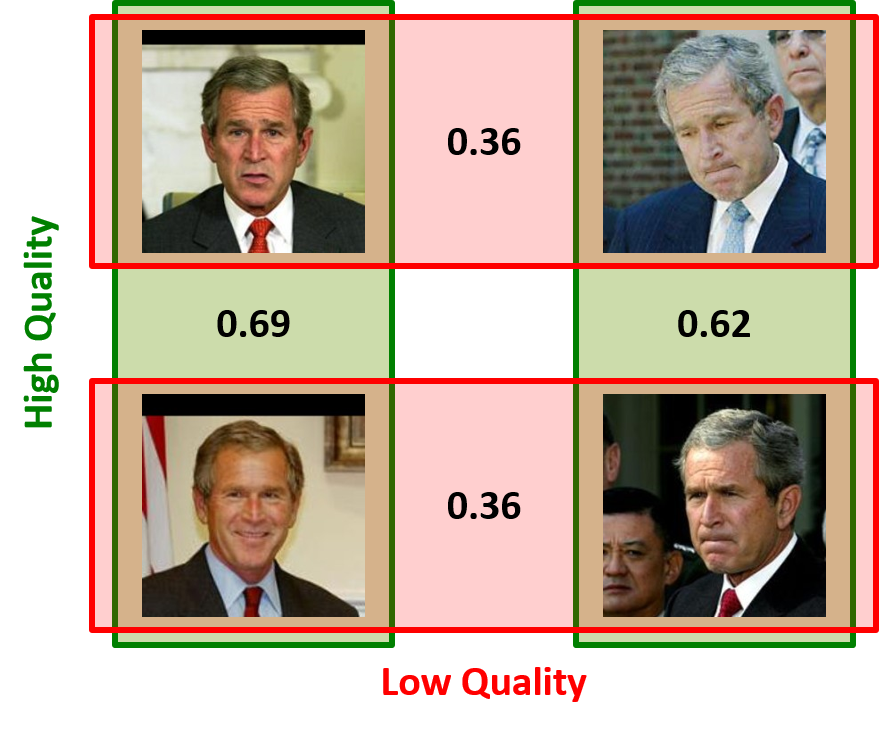}\hspace{5mm}\vspace{-3mm}
\caption{Illustration of the pairwise quality issue. Face images in the left and right columns are individually of high and low qualities, respectively. However, when compared with each other, they can produce both high (good) and low (bad) genuine similarity scores. Similarity scores are from COTS-A with range of $[0,1]$.
}
\label{fig:exGWBush}
\end{figure}

\section{Automatic Prediction of Face Quality}
Given that we have obtained face image quality labels for the LFW database, we now wish to train a model to automatically predict the quality of a probe (previously unseen) face image.
Rather than trying to handcraft a set of image features for our task of predicting face image quality, we make use of features extracted from a deep convolutional neural network which was trained for face recognition purposes by Wang \etal \cite{WangOtto}. The features are 320-dimensional, so we refer to them as \emph{Deep-320} features. The deep network in \cite{WangOtto} was trained on the CASIA WebFaces database \cite{YiScratch_CASIA}.
Using the Deep-320 face image features, we train a support vector regression (SVR) \cite{libsvm} model with radial basis kernel function (RBF) to predict either the normalized comparison scores ($z_{ij}$) from commercial matchers or the human quality ratings. The parameters for SVR (cost, epsilon, and gamma for RBF) are determined via grid search on a validation set of face images.
\iftoggle{notes}{
\textcolor{blue}{[Good to give two expressions for predictor here, one that was trained on labels from the crowd and the other based on scores from the COTS matcher.]}
}




\section{Experimental Evaluation}
As stated earlier, the aim of this work is twofold:
\begin{enumerate}
\item Establish the target, or ``ground truth", quality values of a face image database.
\item Use the quality-labeled face image database to train a model to predict the target quality values using features automatically extracted from an unseen test face image (prior to matching).
\end{enumerate}
Hence, in Sec.~\ref{sec:eval_target}, we first evaluate the \emph{target} face quality values to determine their utility for automatic recognition. In Sec.~\ref{sec:eval_predicted_lfw} we then evaluate how well the target quality values can be predicted by the proposed model for automatic face image quality on the LFW database, and in Sec.~\ref{sec:eval_predicted_ijba} we evaluate the utility of the proposed face image quality values for recognition of face images and video frames from the IJB-A database \cite{KlareIJBA}.

Following the methodology advocated by Grother and Tabassi \cite{GrotherPAMI2007}, we evaluate the face quality measures using the \emph{Error versus Reject (EvR) curve} which evaluates how efficiently rejection of low quality samples results in decreased error rates. The EvR curve plots an error rate (FNMR or FMR) versus the fraction of images removed/rejected, where the error rates are re-computed using a fixed threshold (\eg overall FMR = 0.01\%) after a fraction of the images have been removed. 
We additionally evaluate the utility of the proposed face image quality predictors for improving template-based matching in the IJB-A protocol \cite{KlareIJBA} and provide visual inspections of face images rank-ordered by the proposed face image quality predictor.




\begin{figure}[!t]
\centering
\subfloat[FNMR]{\includegraphics[width=0.8\linewidth]{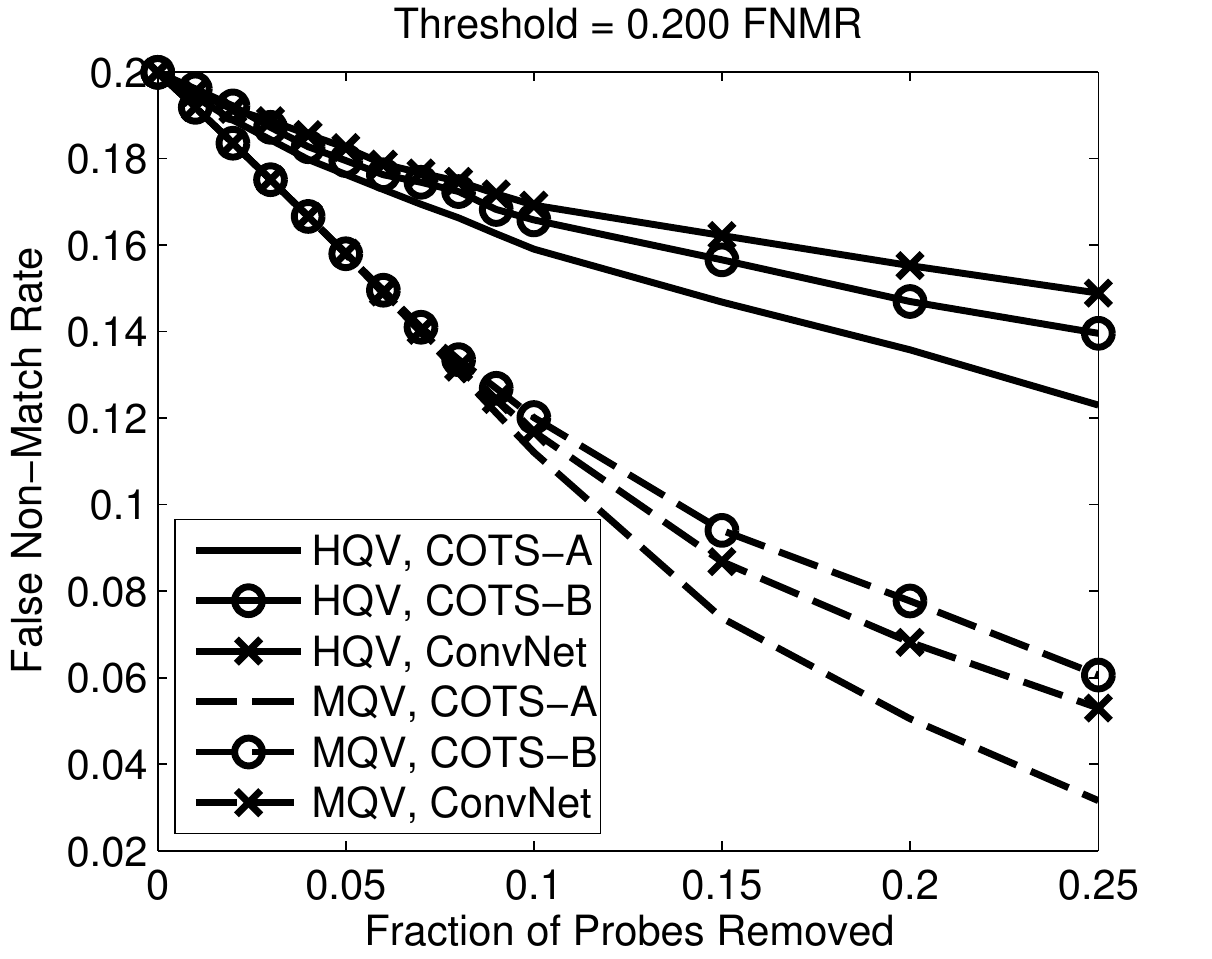}%
\label{fig_actual_fnmr}}
\hfill
\subfloat[FMR]{\includegraphics[width=0.8\linewidth]{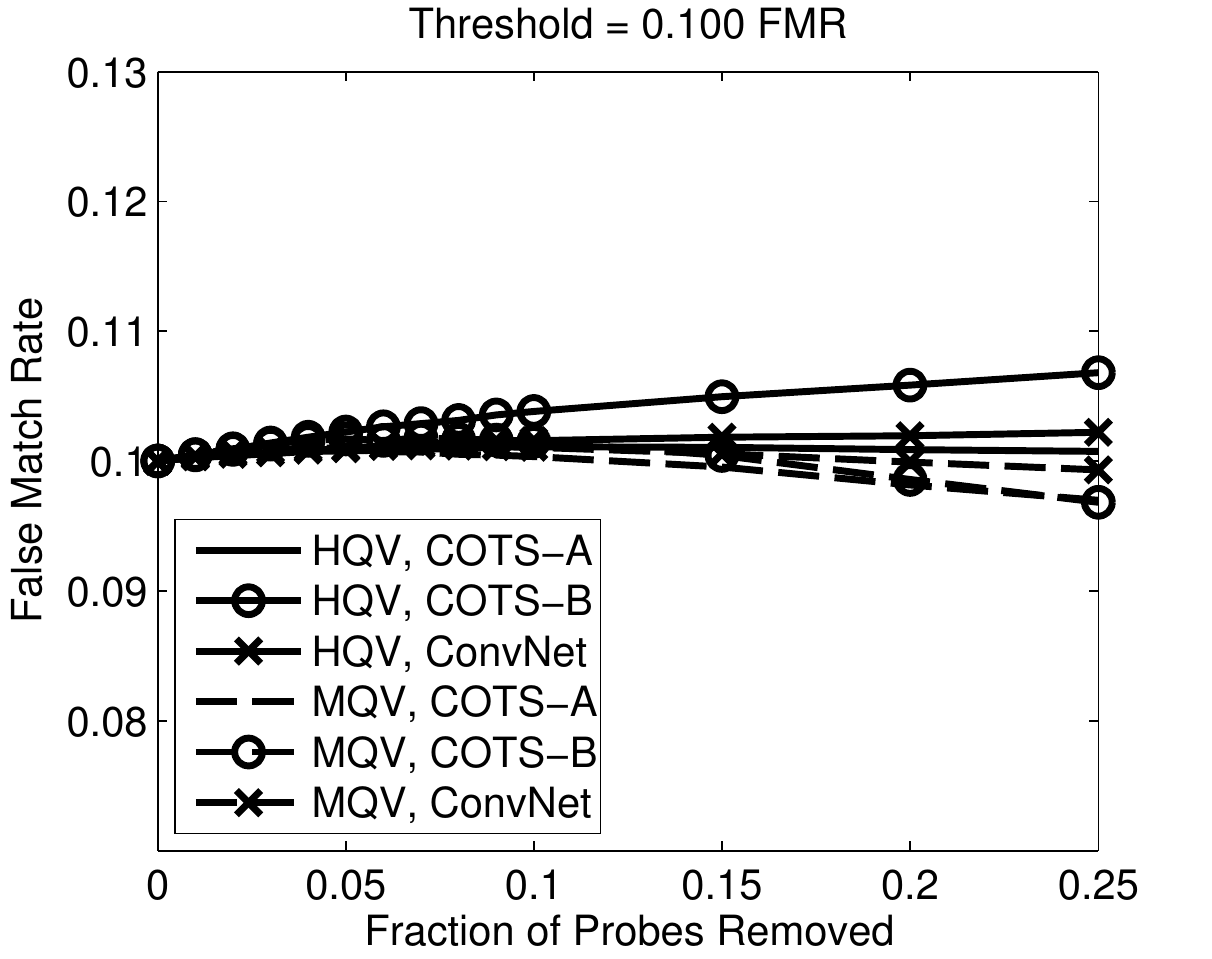}%
\label{fig_actual_fmr}}
\caption{Error vs. Reject curves for (a) FNMR and (b) FMR on the LFW database (gallery size of 5,749 face images and 7,484 probe face images from LFW \cite{LFWTech}). Probe images were rejected in order of \emph{target} (\ie ``target") quality values of human quality ratings (HQV) or score-based quality values (MQV). Thresholds are fixed at (a) 0.20 FNMR and (b) 0.10 FMR for comparison of the three face matchers (COTS-A, COTS-B, and ConvNet \cite{WangOtto}).}
\label{fig:actual_evr}
\end{figure}

\begin{figure*}[t]
\centering
\begin{minipage}{0.23\linewidth}\centering
\includegraphics[width=\linewidth]{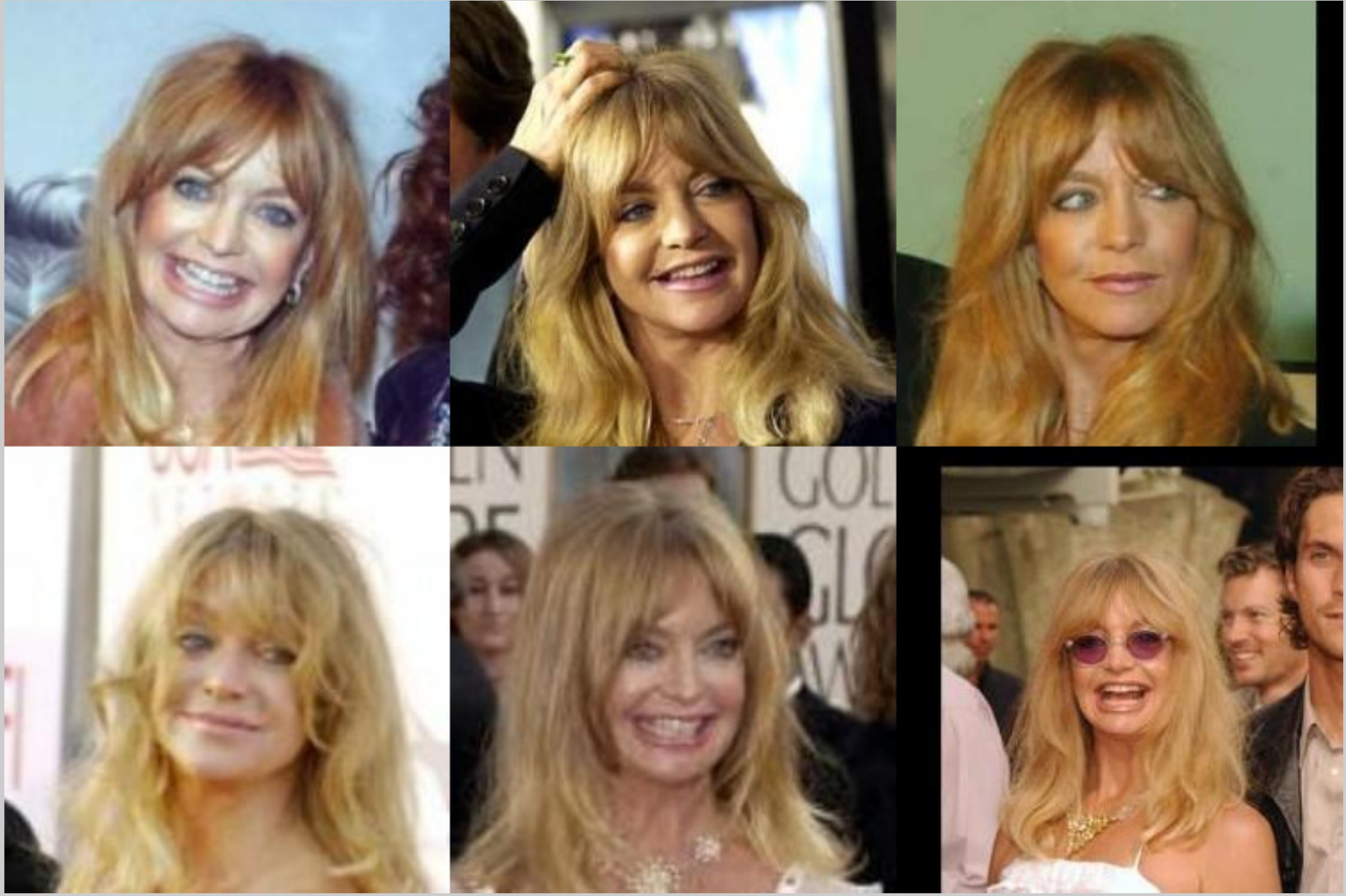}\\ \vspace{3mm}
\includegraphics[width=\linewidth]{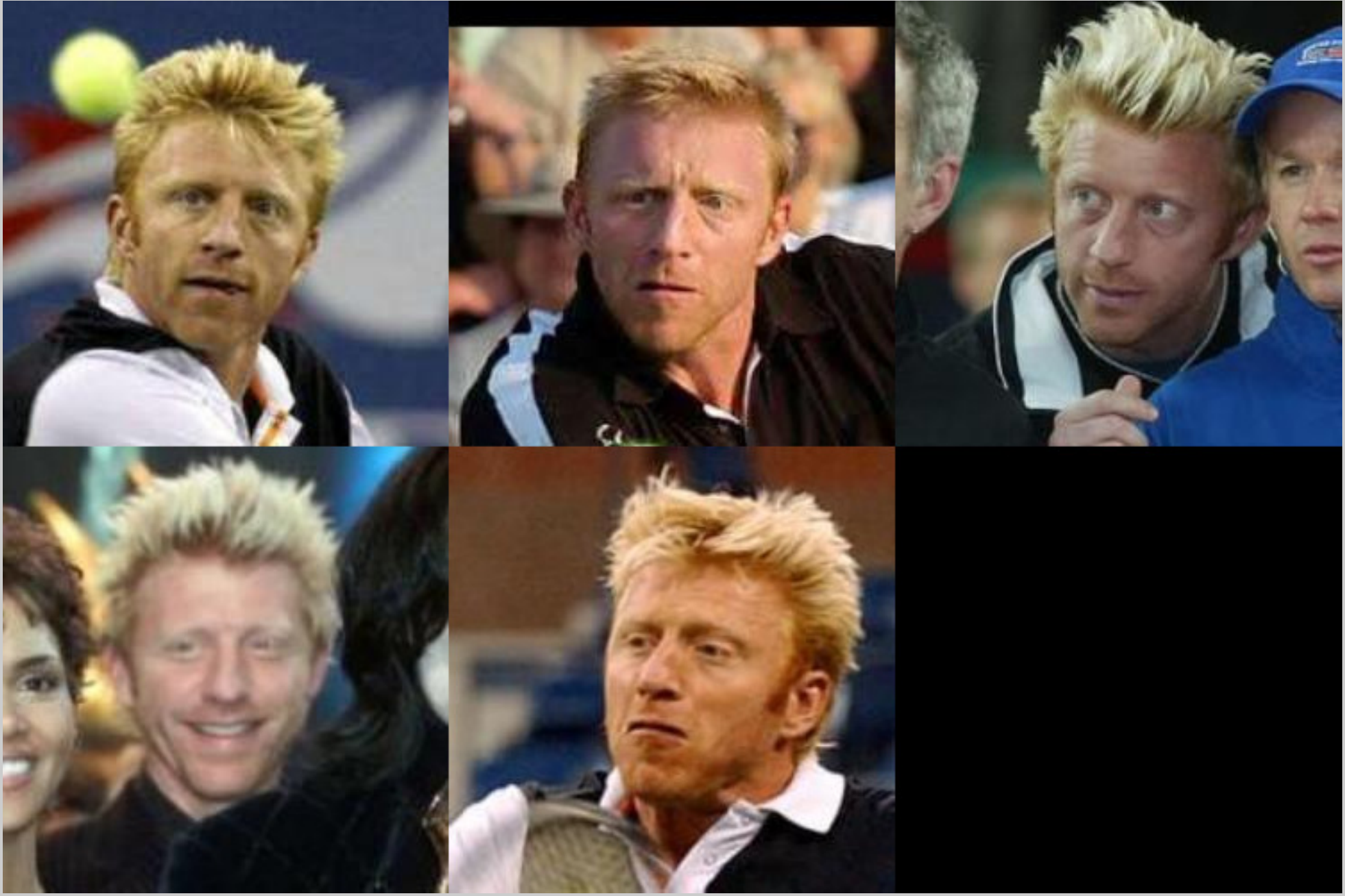}
\end{minipage} \hspace{3mm}
\begin{minipage}{0.4\linewidth}\centering
\includegraphics[width=\linewidth]{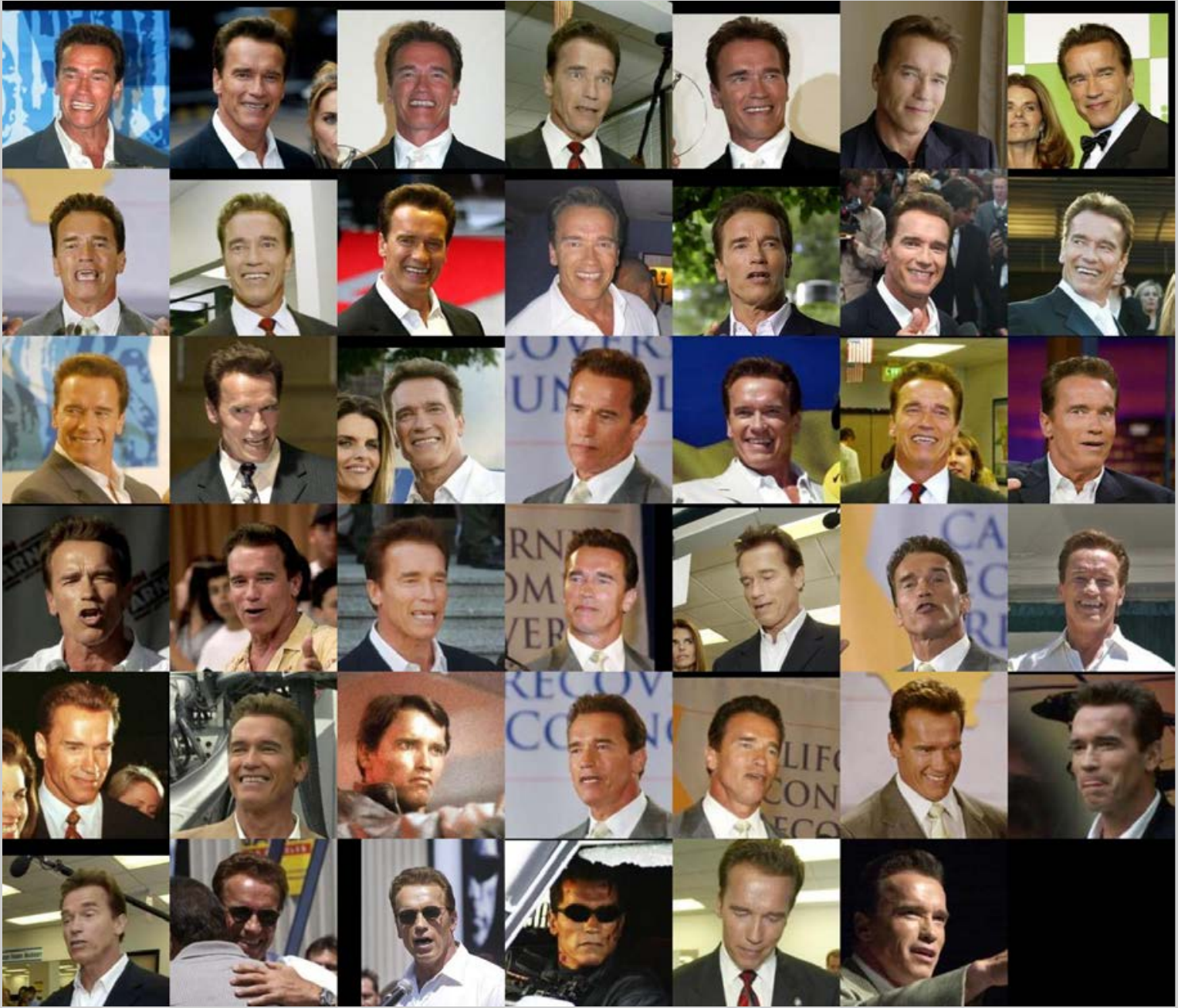}
\end{minipage} \hspace{3mm}
\begin{minipage}{0.3\linewidth}\centering
\includegraphics[width=\linewidth]{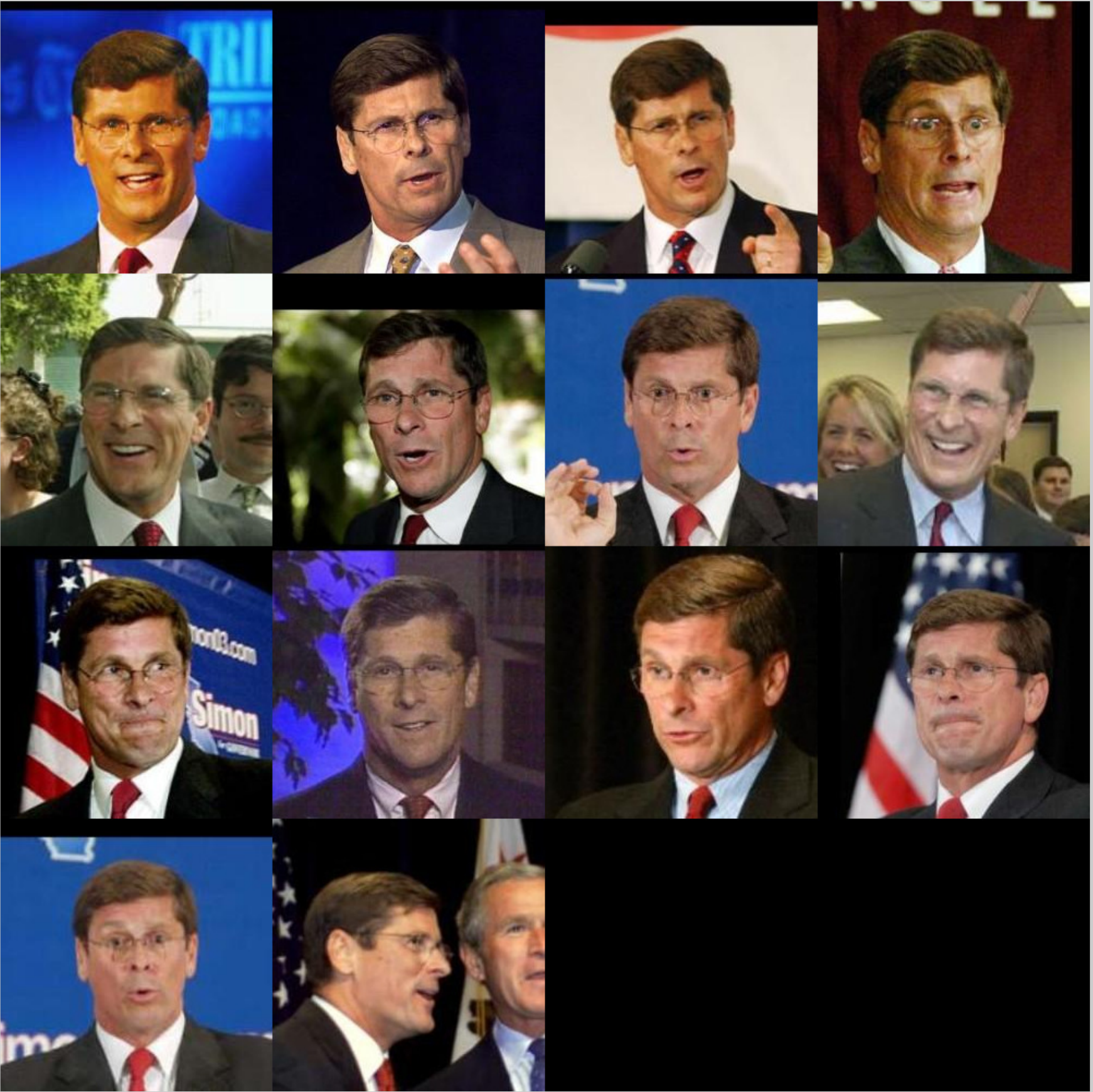} 
\end{minipage}
\caption{Face images of four subjects from LFW \cite{LFWTech} rank-ordered by the predicted human quality ratings from the proposed HQV method. Face images are shown in order of decreasing face quality. For each of the four example subjects, the Spearman correlation between the target and predicted rank orderings are 0.94, 0.90, 0.72, and 0.50.}
\label{fig:predictLFWgoodHuman}
\end{figure*}

\subsection{Target Face Image Quality Values}
\label{sec:eval_target}
Face images in the LFW database are ``ground-truth" labeled using the methods discussed in Section~\ref{sec:qualityLabels}. 
We refer to these as \emph{target} quality values, where score-based quality values are denoted as Matcher Quality Values (MQV) and human quality ratings are denoted as Human Quality Values (HQV). 
For sake of comparison, we use only the 7,484 probe images from LFW for both MQV and HQV methods, and evaluate both MQV and HQV using the gallery and probe setup of the LFW database detailed in Section~\ref{sec:mqv}.

Fig.~\ref{fig:actual_evr} plots EvR curves for both target quality values (MQV and HQV), evaluated for three different face matchers (COTS-A, COTS-B, and ConvNet \cite{WangOtto}). Because the matchers are of different strengths, a common initial FNMR and FMR of 0.20 and 0.10, respectively, were chosen for the evaluation of all three matchers. Fig.~\ref{fig_actual_fnmr} shows that removing low-quality probe images in order of HQV decreases FNMR for all three matchers, indicating that human quality ratings are correlated with recognition performance.
However, MQV is much more efficient in reducing FNMR. This is expected because the score-based target quality values are computed from the same comparison scores used to compute the FNMR for each matcher. Again, the score-based target quality values serve as an ``oracle" for a desirable quality measure. 

The utility of the target quality values in terms of reducing FMR in Fig.~\ref{fig_actual_fmr} is not as apparent; in fact, removing low-quality images based on HQV clearly \emph{increases} FMR for COTS-B, though the magnitude of the increase is small (removing 25\% of the probe images increases FMR by 0.14\%). The relation between face quality and impostor scores (\ie FMR) is generally less of a concern. For biometric quality, in general, we desire \emph{high} quality samples to produce \emph{low} impostor similarity scores, but \emph{low} quality samples may also produce \emph{low} (or even lower) impostor scores. If this is the case, low quality face images may be beneficial to FMR for empirical evaluation, but still undesirable operationally. Due to this conundrum, we focus on the effect of face quality on FNMR for the remainder of the experiments.



\subsection{Predicted Face Image Quality Values}
\label{sec:eval_predicted}
The proposed framework for automatic prediction of face image quality (using both human ratings and score-based quality values as targets) is used to predict the quality of face images from the LFW \cite{LFWTech} and IJB-A \cite{KlareIJBA} databases. The prediction models for both databases are trained using LFW face images, and the experimental protocols are detailed in the following sections. 

\begin{table}[t]
\centering
\footnotesize
\renewcommand{\arraystretch}{1.3}
\caption[caption]{Spearman rank correlation (mean $\pm$ standard deviation over 10 random splits of LFW images) between target and predicted Matcher Quality Values (MQV) and Human Quality Values (HQV) }
\label{tab:corrPrediction} 
\vspace{-3mm}
\begin{tabular}{|c||c|c|c|}
\hline
& \multicolumn{3}{c|}{\textbf{Matcher}}\\ \cline{2-4}
 & COTS-A & COTS-B & ConvNet\\ 
\hline
\textbf{MQV} & 0.558 $\pm$ 0.023 &  0.442 $\pm$ 0.026 & 0.459 $\pm$ 0.022\\ \hline
\textbf{HQV} &  \multicolumn{3}{c|}{0.585 $\pm$ 0.019} \\ \hline
\end{tabular}
\end{table}

\begin{figure*}[t]
\centering
\subfloat[COTS-A, $z_{ij}$]{\includegraphics[width=0.33\linewidth]{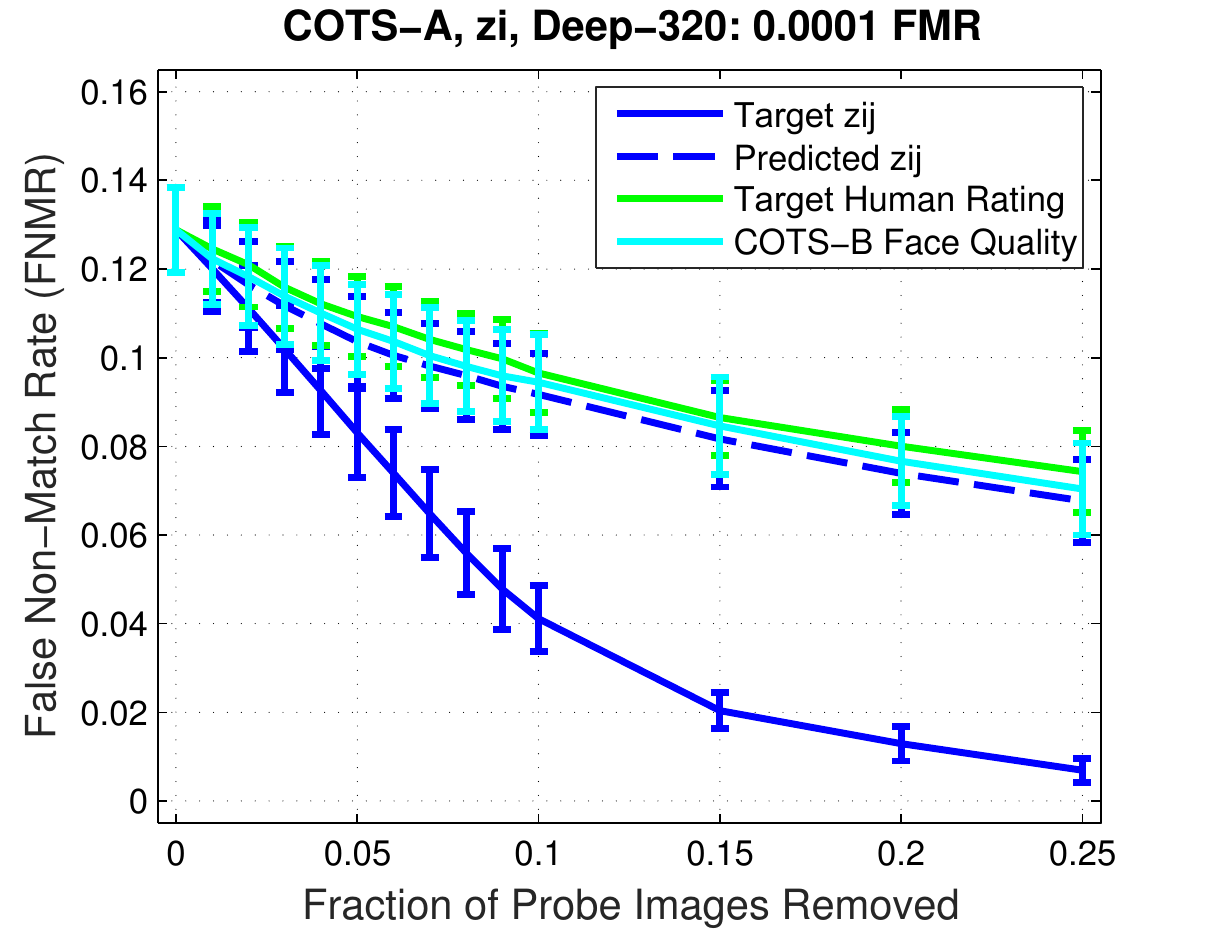}%
\label{fig_first_case}}
\hfil
\subfloat[COTS-A, Human]{\includegraphics[width=0.33\linewidth]{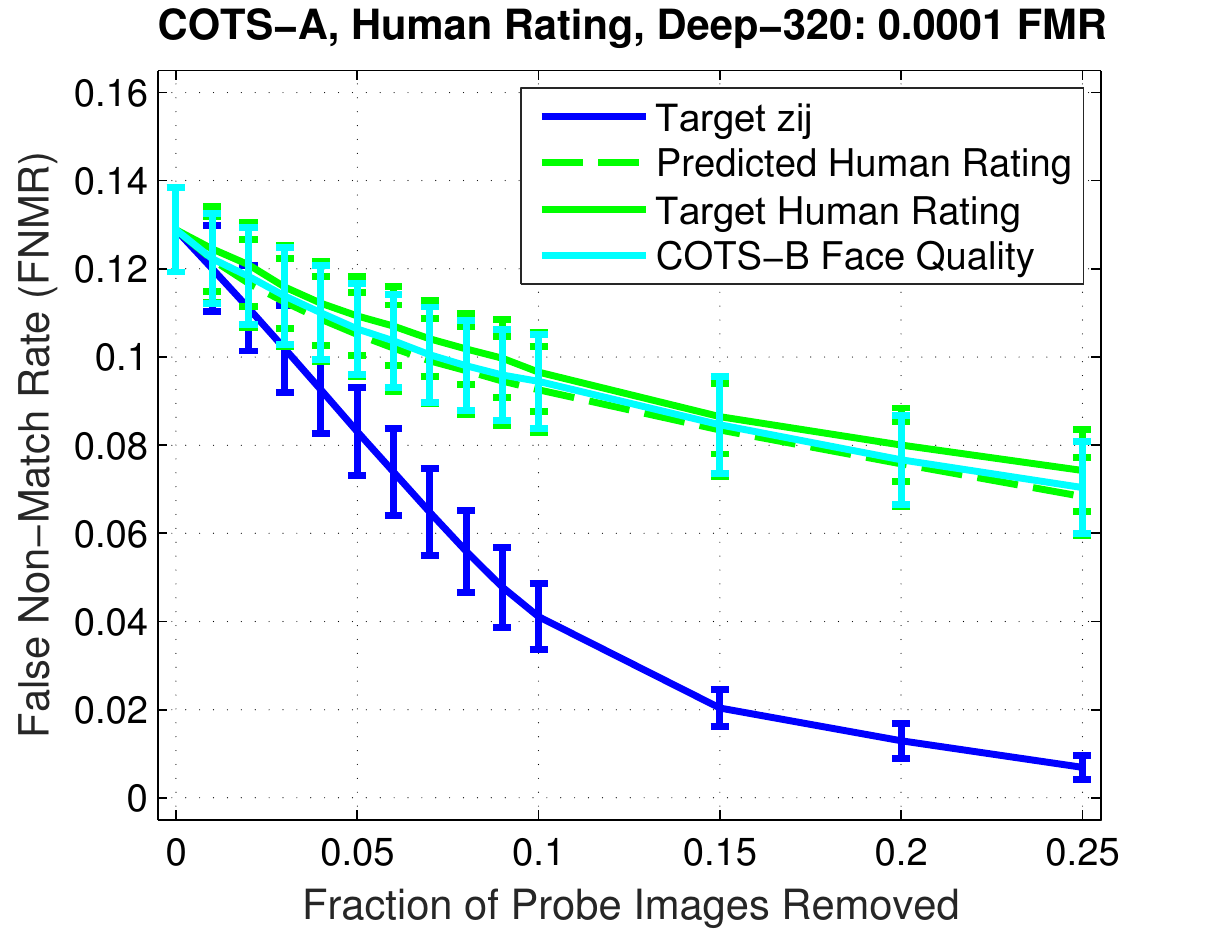}%
\label{fig_second_case}}
\hfil
\subfloat[COTS-B, $z_{ij}$]{\includegraphics[width=0.33\linewidth]{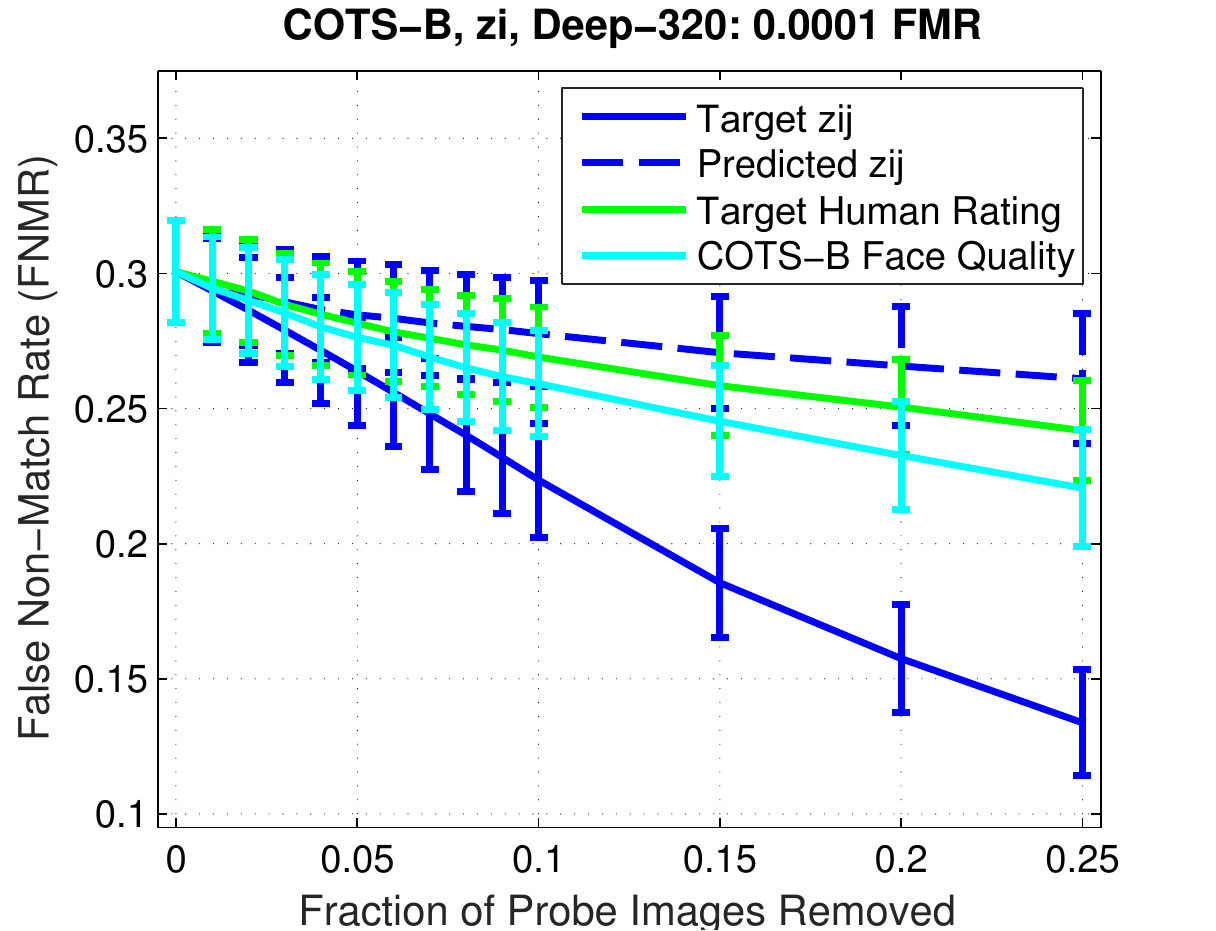}%
\label{fig_third_case}}
\caption{Error vs. Reject curves for target and predicted face image quality values (MQV and HQV) for the LFW database. The curves show the efficiency of rejecting low quality face images in reducing FNMR at a fixed FMR of 0.01\%. The models used for the face quality predictions in (a)-(c) are support vector regression on the deep-320 features from ConvNet in \cite{WangOtto}. 
}
\label{fig:evrLFW}
\end{figure*}

\subsubsection{Train, Validate, and Test on LFW}
\label{sec:eval_predicted_lfw}
We first divide 7,484 face images of the 1,680 subjects with two or more images in LFW into 10 random splits for training and testing, where ${2}/{3}$ and ${1}/{3}$ of the subjects are randomly split into training and testing sets, respectively. For each split, we then conduct 5-fold cross-validation within the training set to tune the parameters for the support vector regression model via grid search. The selected set of parameters is applied to the full training set to result in a single model for each of the 10 splits, which are then used to predict the quality labels of the images in each of the 10 test sets. This framework ensures subject-disjoint training and testing sets, and parameter selection is conducted within a validation set,  not optimized for the test sets. 

Table~\ref{tab:corrPrediction} gives the rank correlation (mean and standard deviation over the 10 splits) between the target and predicted quality values for human quality ratings and score-based quality values for (MQV separately for COTS-A, COTS-B, and ConvNet \cite{WangOtto} matchers). We observe that prediction of human quality ratings is more accurate than prediction of score-based quality for all three matchers, likely due to the difficulty in predicting particular nuances of each matcher. Fig.~\ref{fig:predictLFWgoodHuman} shows images sorted by predicted HQV of four example subjects from LFW with \emph{strong} rank correlation (Spearman) between target and predicted human quality values.


To evaluate the quality values in the context of automatic face recognition performance, error vs. reject curves (for FNMR at fixed 0.01\% FMR) are plotted in Fig.~\ref{fig:evrLFW} for both target and predicted quality values (MQV and HQV). The figures demonstrate that rejecting low quality face images based on predicted $z_{ij}$, predicted human ratings, or the COTS-B measure of face quality, results in comparable efficiency in reducing FNMR (\eg removal of 5\% of probe images lowers FNMR by {$\sim$2\%}). However, none of the methods are near as efficient as rejecting images based on the target $z_{ij}$ values, which serve as an oracle for a predicted face quality measure that is highly correlated with the recognition performance.

\subsubsection{Train and Validate on LFW, Test on IJB-A}
\label{sec:eval_predicted_ijba}
In this framework, we conduct 5-fold cross-validation over the 7,484 LFW images (folds are subject-disjoint) to determine the parameters for the support vector regression model via grid search. We then apply the selected set of parameters to all of the LFW training images. This model trained on the LFW database is then used to predict the quality of face images in the IJB-A database \cite{KlareIJBA}. 

\iftoggle{notes}{
\textcolor{blue}{[We do not do any additional training using the ``training'' images defined from the IJB-A protocol...]}
}

For evaluation on the IJB-A database, we follow the template-based matching ``Compare" (\ie verification) protocol \cite{KlareIJBA}, which consists of 10 random splits (bootstrap samples) of the 500 total IJB-A subjects. For each split, 333 subjects are randomly sampled for training and the remaining 167 subjects for testing. However, note that we do not actually do any training with IJB-A images; our face quality models are trained on LFW.  
In template-based matching, multiple face images and/or video frames are available for a subject in the gallery and/or probe sets. Baseline results for score-level fusion (SLF) using \emph{max} and \emph{mean} rules are given are given in Figure~\ref{subfig:ijba_ROC} for the COTS-A and ConvNet matchers. COTS-B was not used for evaluation on IJB-A database because of a much higher failure to enroll (FTE) rate than COTS-A and ConvNet matchers. Figure~\ref{subfig:ijba_ROC} shows that mean fusion is slightly better than max fusion for both matchers, and that COTS-A performs better than ConvNet matcher at lower FMR. At 1\% FMR, COTS-A and ConvNet are comparable; FNMR is 42.0\% and 47.7\% for COTS-A and ConvNet, respectively (mean fusion).	



\begin{figure*}
\centering
\includegraphics[width=0.9\linewidth,trim={1.5cm 5.25cm 1.5cm 5cm},clip]{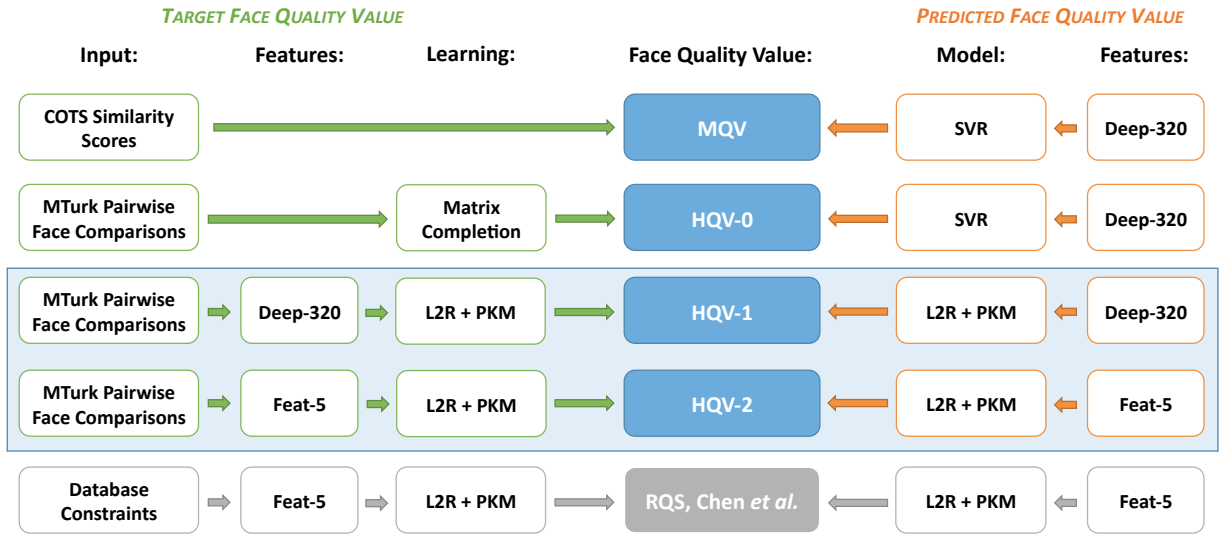}
\caption{Flowcharts indicating the components of defining target face quality values (left side) and then predicting the target face quality values (right side) for the five methods evaluated on the IJB-A database \cite{KlareIJBA}. MQV and HQV-0 are the methods proposed in this paper, while RQS is proposed by Chen \emph{et al.} \cite{Chen2015}. HQV-1 and HQV-2 are variants of HQV-0 with some components of RQS plugged in to evaluate the impact of the input used to define target quality values, the image features (Deep-320 vs. Feat-5), and the learning framework used (support vector regression (SVR) vs. learning to rank with polynomial kernel mapping function (L2R $+$ PKM)). }
\label{fig:targetPredictedMethods}
\end{figure*}

\begin{figure*}
\centering
\subfloat[COTS-A and ConvNet]{\label{subfig:ijba_ROC}
\includegraphics[width=0.33\linewidth,trim={0 0 6mm 6mm},clip]{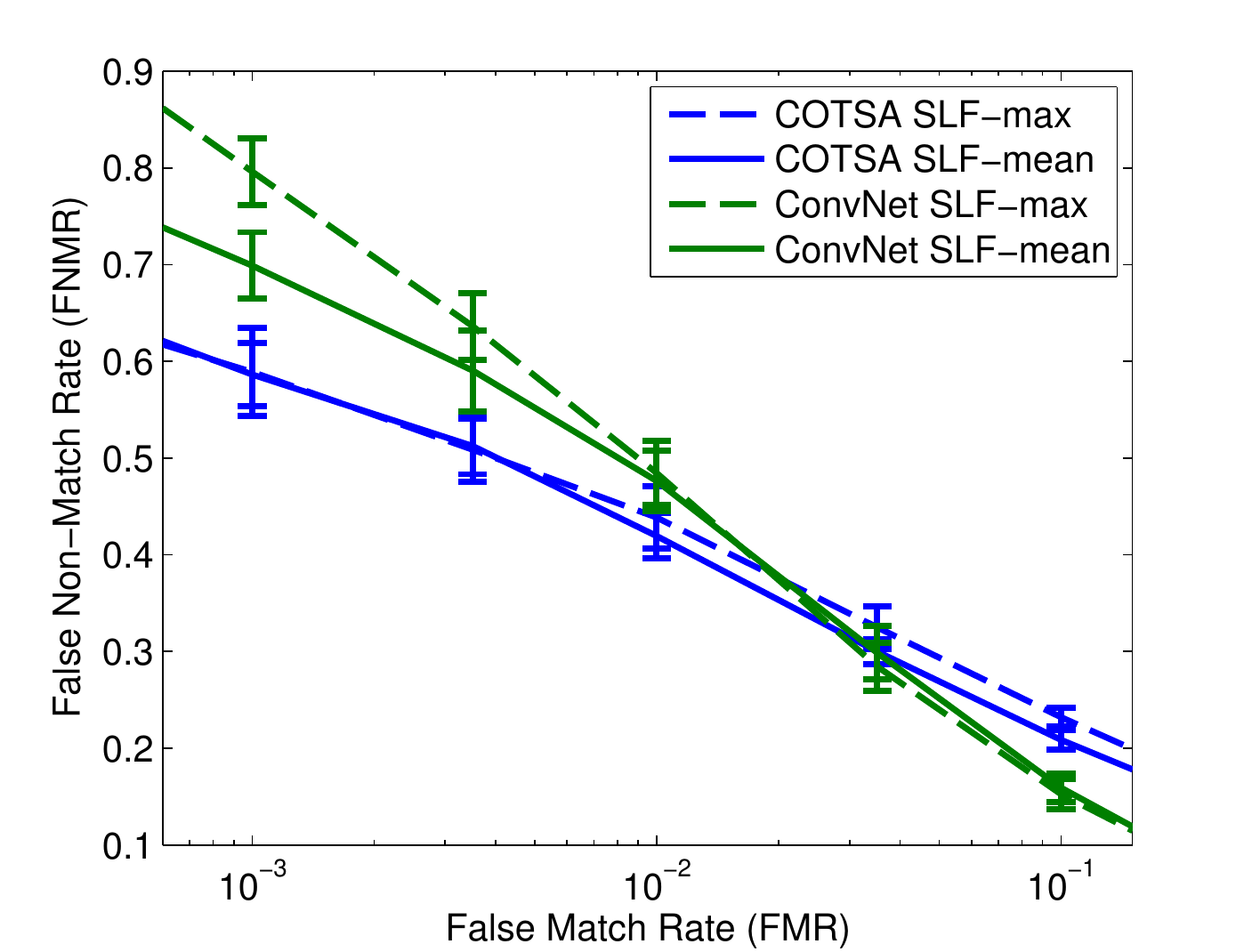}
}
\subfloat[COTS-A]{\label{subfig:ijba_COTSA}
\includegraphics[width=0.33\linewidth,trim={0 0 6mm 6mm},clip]{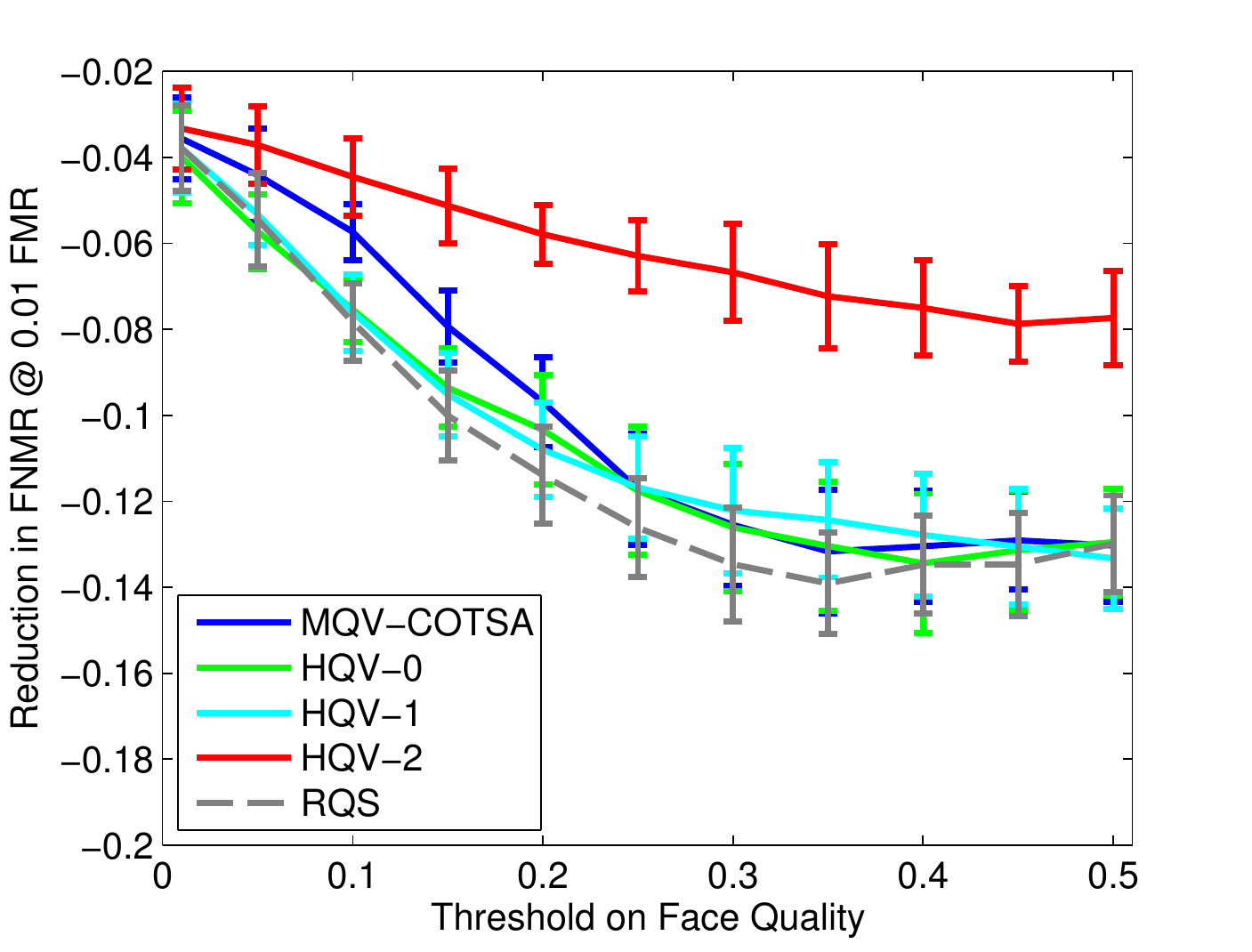}
}
\subfloat[ConvNet]{\label{subfig:ijba_ConvNet}
\includegraphics[width=0.33\linewidth,trim={0 0 6mm 6mm},clip]{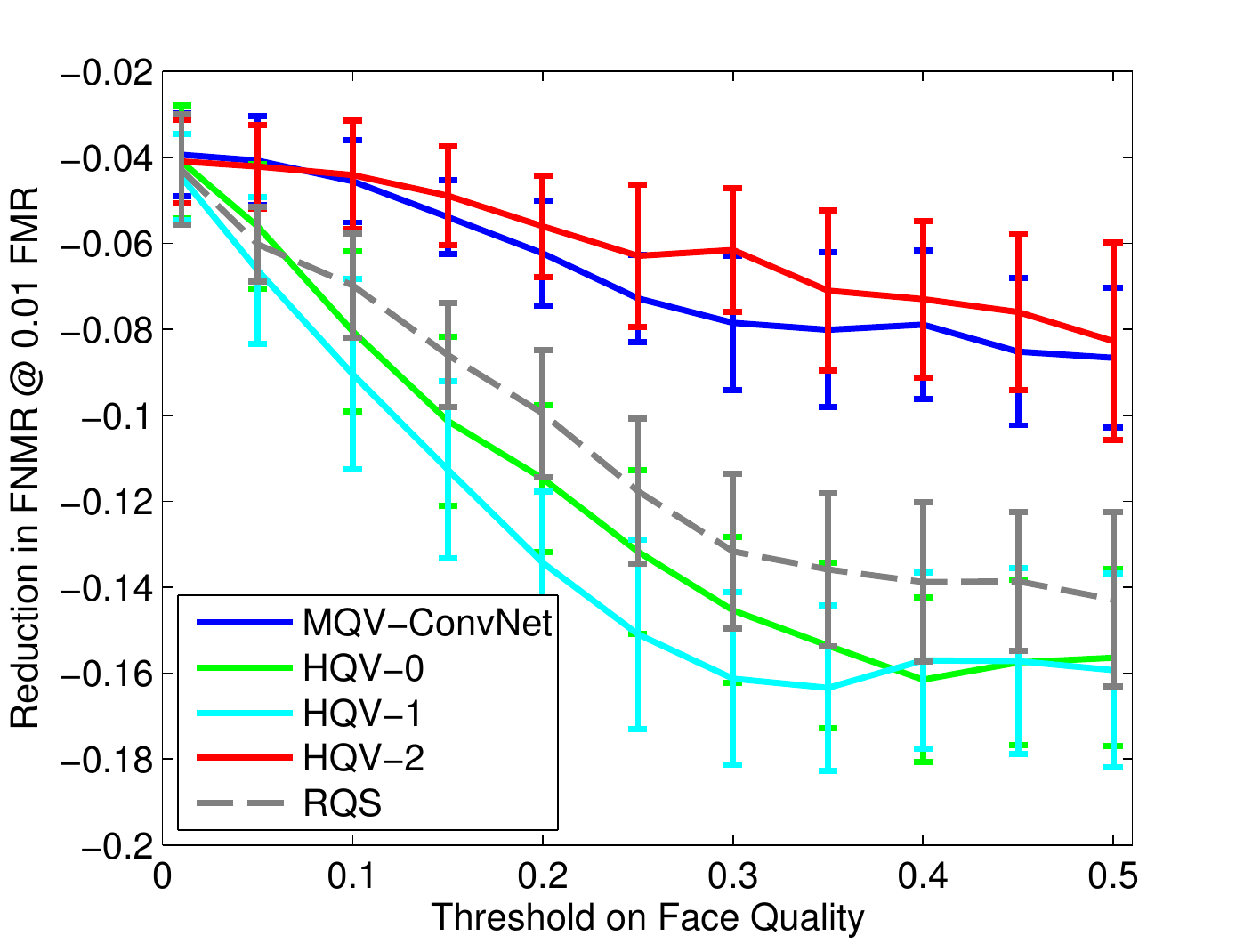}
}
\caption{Results for the compare (\ie verification) protocol of the IJB-A database \cite{KlareIJBA}. All curves in (a)-(c) show mean performance and error bars give standard deviation in performance over the 10 splits in the protocol. (a) Receiver Operating Characteristic (ROC) for COTS-A and ConvNet \cite{WangOtto} matchers, where score-level fusion (SLF) is applied to the multiple face samples per subject for template-based matching of the IJB-A protocol. Using thresholds on face image quality measures to determine which face samples in a template to use for matching, (b) and (c) plot reduction in FNMR at 1\% FMR, showing that FNMR decreases as the face quality thresholds are increased. Flowcharts providing details of each face quality method (MQV, HQV-0, etc.) are given in Fig.~\ref{fig:targetPredictedMethods}. The RQS method is proposed by Chen \emph{et al.} \cite{Chen2015}.} 
\label{fig:ijba}
\end{figure*}

For the IJB-A database, we compare the proposed MQV and HQV methods with Chen \etal's Rank-based Quality Score (RQS) \cite{Chen2015}. The RQS method defines pairwise constraints on face images based on a relative ordering of face image databases. The learning to rank (L2R) framework of Parikh and Grauman \cite{ParikhGrauman_ICCV2011} is used to learn five different ranking functions, one for each of five different image features (HoG, Gabor, Gist, LBP, and CNN features), which we refer to as \emph{Feat-5}. The five ranking functions are then combined with a polynomial kernel mapping (PKM) function. To predict the RQS of a new test image, Feat-5 image features are extracted and multiplied by the weight vector obtained from the (L2R $+$ PKM) framework. 

Using the RQS\footnote{http://jschenthu.weebly.com/projects.html} and the L2R\footnote{\url{https://filebox.ece.vt.edu/~parikh/relative.html}} codes, both made publicly available by their authors, we combine different components of the RQS method with the human pairwise comparisons from MTurk and the Deep-320 features to evaluate the impact of these components. Figure~\ref{fig:targetPredictedMethods} shows a flowchart of the variants of the proposed HQV method (HQV-0, HQV-1, and HQV-2), where MTurk pairwise comparisons are the input to establish target quality values, but Feat-5 features and/or (L2R $+$ PKM) framework are used instead of Deep-320 features and/or matrix completion. The flowcharts for the proposed MQV method and Chen \etal's RQS method are also given in Fig.~\ref{fig:targetPredictedMethods}; in total, we evaluate five different face quality methods. 


Finally, to evaluate the utility of face quality values to recognition performance, we incorporate the face quality into template-based matching as follows: given a threshold on the face quality, the template for a subject consists of only the faces with quality at least as high as the threshold; if there are no faces with quality above the threshold, select only the single best face. Score-level mean fusion is then applied to the scores from the selected faces.
 
Figures~\ref{subfig:ijba_COTSA} and \ref{subfig:ijba_ConvNet} report the reduction in FNMR at fixed 1\% FMR when the threshold on face quality is varied; the thresholds considered are $n/100$ where $n$ is the $n$th percentile of the face quality values for all images and videos in the given testing split of IJB-A database. This evaluation is similar to the Error vs. Reject (EvR) curve except that the number of scores used to compute performance remains the same as face samples are removed.  Because the face quality methods MQV, HQV, and RQS each use their own face detector, the face quality for images in which \emph{any} of the detectors failed are all set to the lowest quality value, so these images are removed first for all face quality methods, providing a fairer comparison.

A few observations can be made from Figs.~\ref{subfig:ijba_COTSA} and \ref{subfig:ijba_ConvNet} about the different face quality methods. 
(i) MQV performs quite well at reducing FNMR for COTS-A, but is much worse for the ConvNet matcher. This may be because the Deep-320 features used for MQV face quality prediction are the same features used by the ConvNet matcher, so the MQV for ConvNet 
(ii) HQV-2 performs poorly, while HQV-1 effectively reduces FNMR for both matchers, suggesting that Deep-320 features are more powerful for predicting the human quality ratings than Feat-5 features. 
(iii) HQV-0 and HQV-1 perform comparably for COTS-A, but HQV-1 performs slightly better for the ConvNet matcher. This suggests that the (L2R + PKM) framework may be somewhat better than matrix completion for establishing the target face quality values from pairwise comparisons. 
(iv) HQV-0 and HQV-1 both perform comparable to the RQS method \cite{Chen2015} for both matchers, and all three face quality methods effectively reduce FNMR by removing low-quality face images or videos from IJB-A templates. Using mean score-level fusion of all faces in the templates as a baseline, FNMR is reduced by $\sim$13\% for COTS-A and $\sim$16\% for ConvNet matchers given a quality threshold of the 40th percentile of the distribution of quality values in the training sets. 
Table~\ref{tab:ijba} summarizes the results on the IJB-A verify protocol \cite{KlareIJBA} for the COTS-A and ConvNet matchers with and without the proposed HQV face quality predictor and compares the performance to previously published results on the IJB-A protocol. Performance is reported as True Accept Rate (TAR) at fixed False Accept Rates (FARs) as the protocol suggests \cite{KlareIJBA}. 
Without HQV, COTS-A and ConvNet matchers are very poor compared with the other face recognition methods in Table~\ref{tab:ijba}, but with HQV the performance is greatly improved. At 1\% FAR, TAR for COTS-A (ConvNet) increases from 58.0\% (52.3\%) for standard score-level fusion to 71.5\% (68.5\%) using the proposed HQV method to adaptively select which faces to use in template-based matching. 

\begin{figure*}
\centering
\hspace{-18mm}
\begin{minipage}[]{0.4\linewidth}\raggedleft
\includegraphics[height=0.825\textwidth]{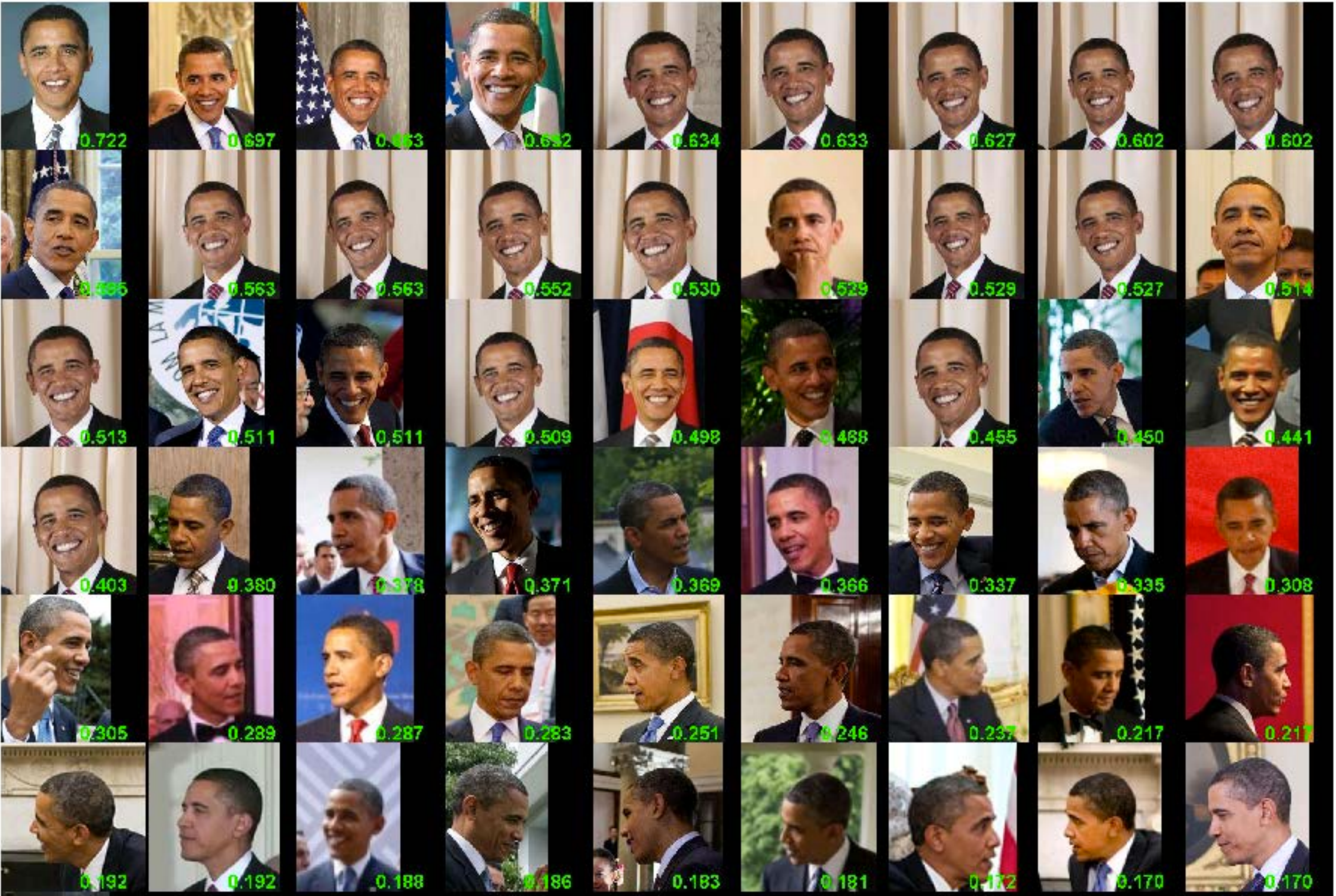}\\\vspace{2mm}
\includegraphics[height=0.825\textwidth]{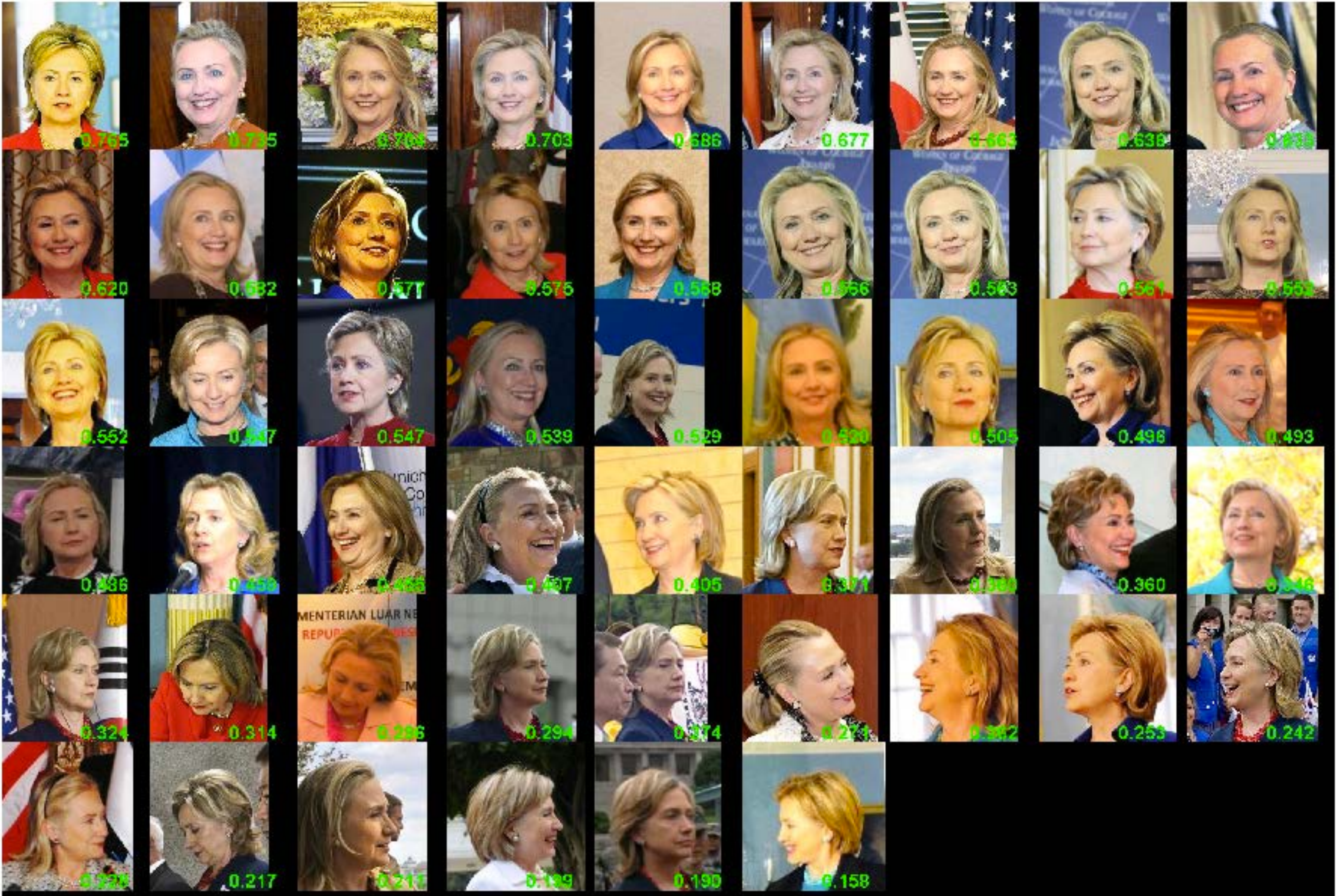}\\\vspace{2mm}
\footnotesize (a) Ranked by the Proposed HQV \hspace{5mm} 
\end{minipage}\hspace{17mm}
\begin{minipage}[]{0.4\linewidth}\raggedright
\includegraphics[height=0.825\textwidth]{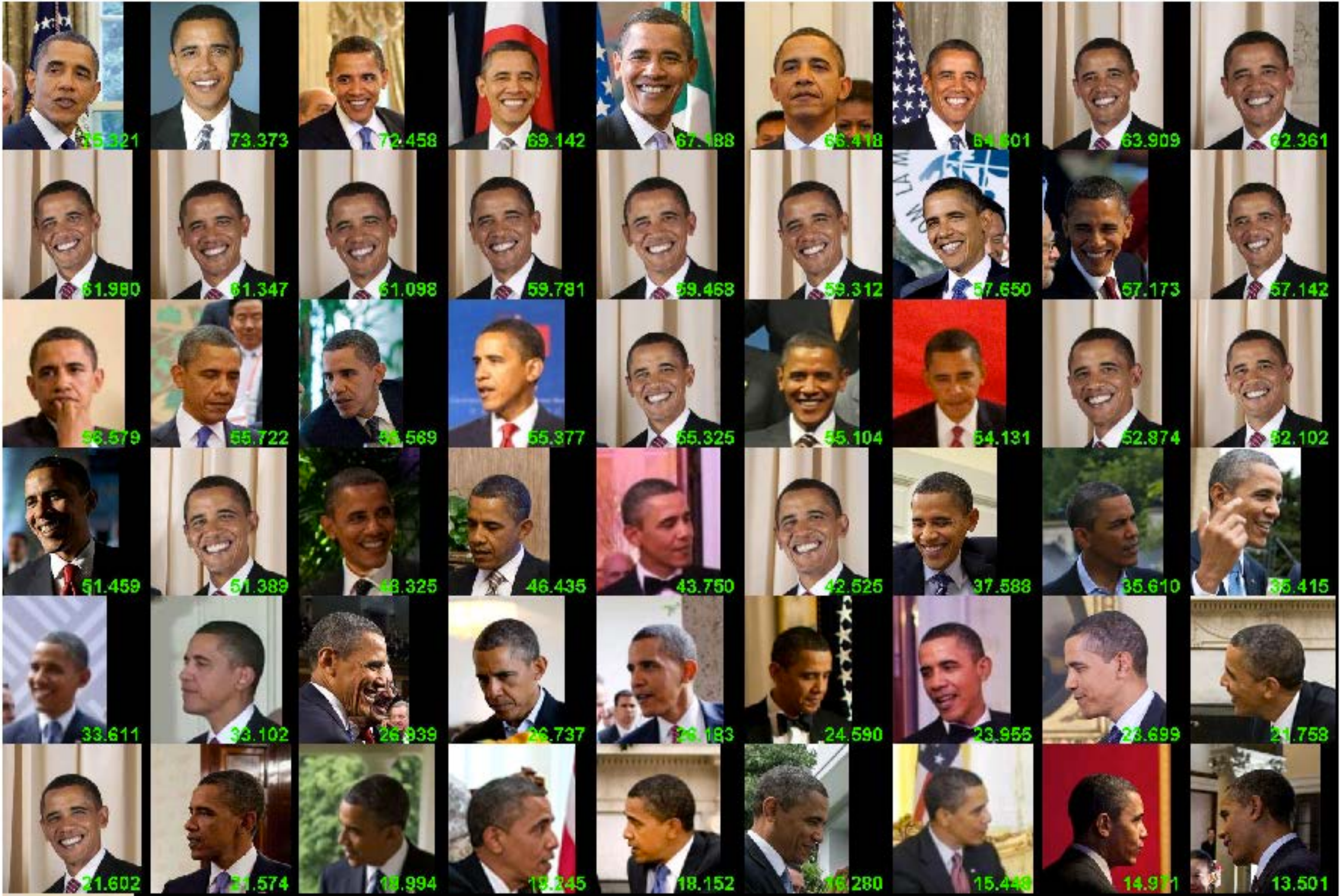}\\\vspace{2mm}
\includegraphics[height=0.825\textwidth]{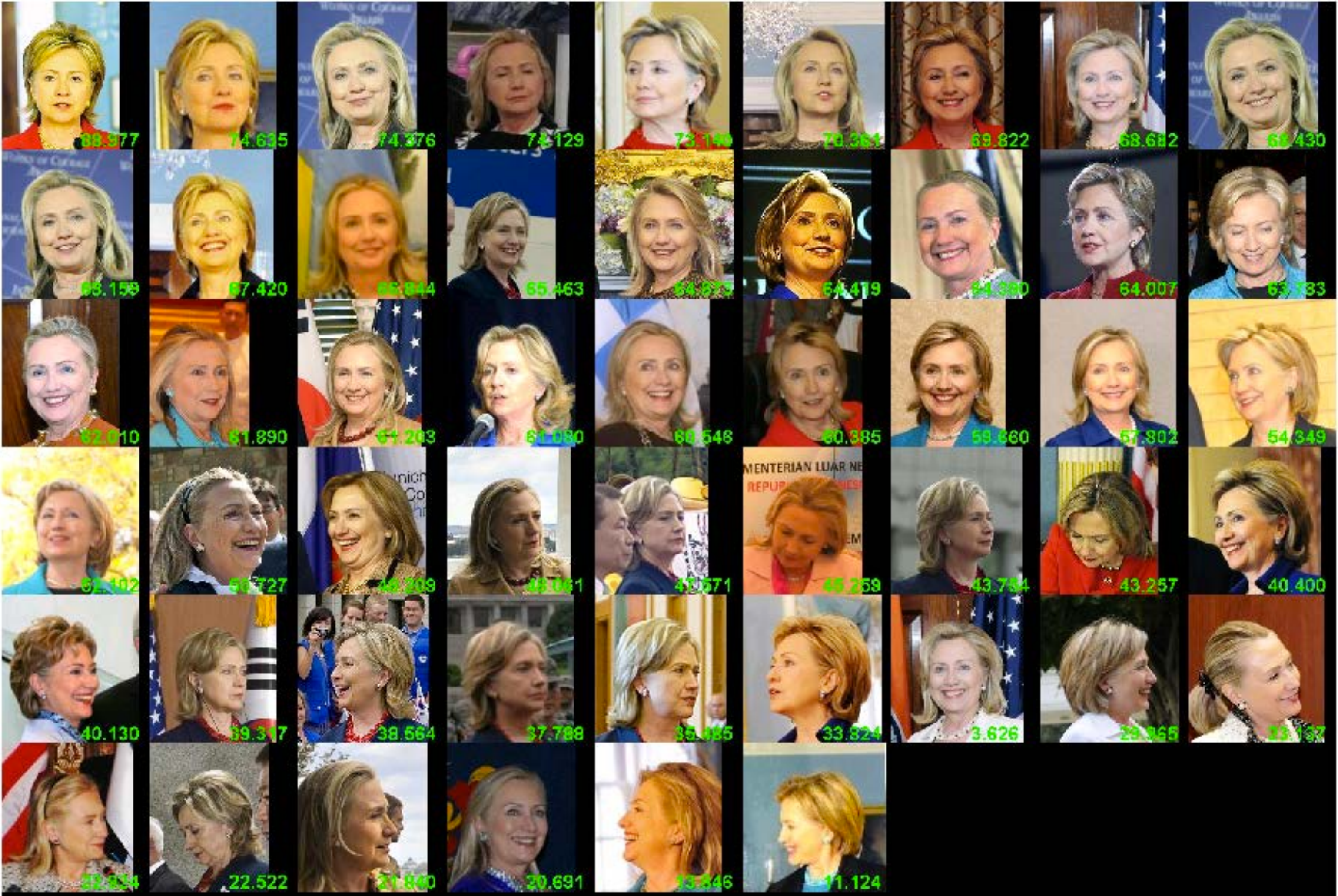}\\\vspace{2mm}
\hspace{25mm} \footnotesize (b) Ranked by RQS \cite{Chen2015}
\end{minipage}
\caption{Face images from two subjects in IJB-A \cite{KlareIJBA} sorted by face image quality (best to worst). The face image qualities were automatically predicted by (a) the proposed approach (SVR model to predict human quality values (HQV) from Deep-320 image features \cite{WangOtto}) and (b) Rank-based Quality Score (RQS) \cite{Chen2015} for comparison.}
\label{fig:ijbaRanked}
\end{figure*}

\begin{figure*}[!t]
\centering
\includegraphics[height=0.11\linewidth]{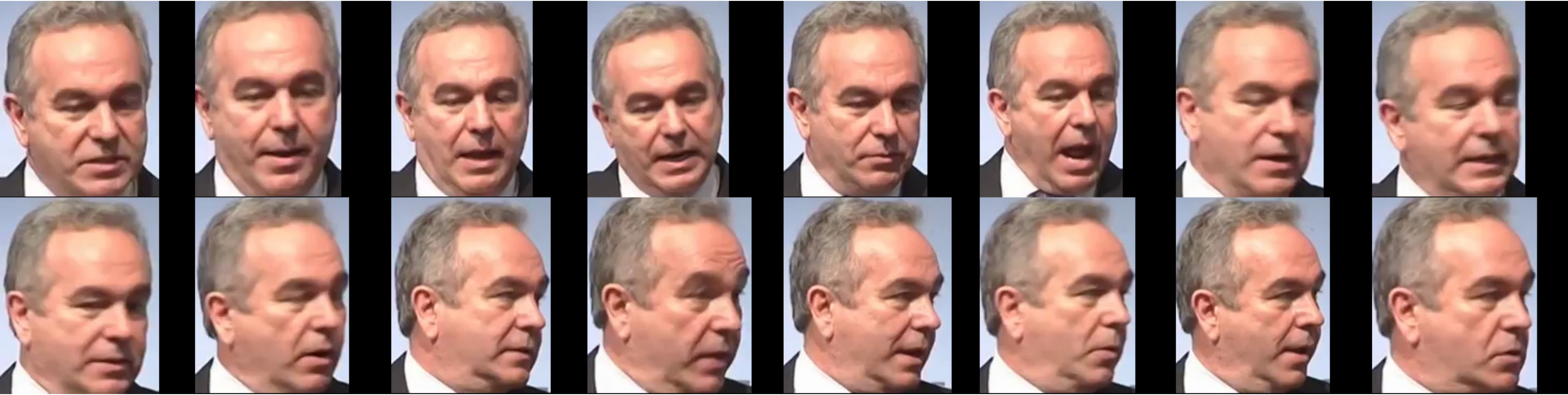}\hspace{3mm}
\includegraphics[height=0.11\linewidth]{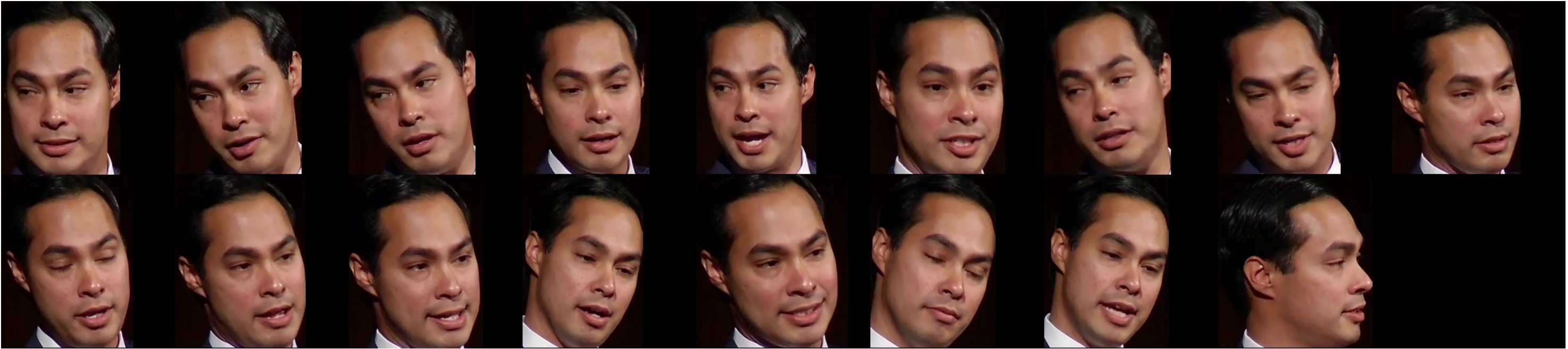}\\\vspace{3mm}
\includegraphics[height=0.18\linewidth]{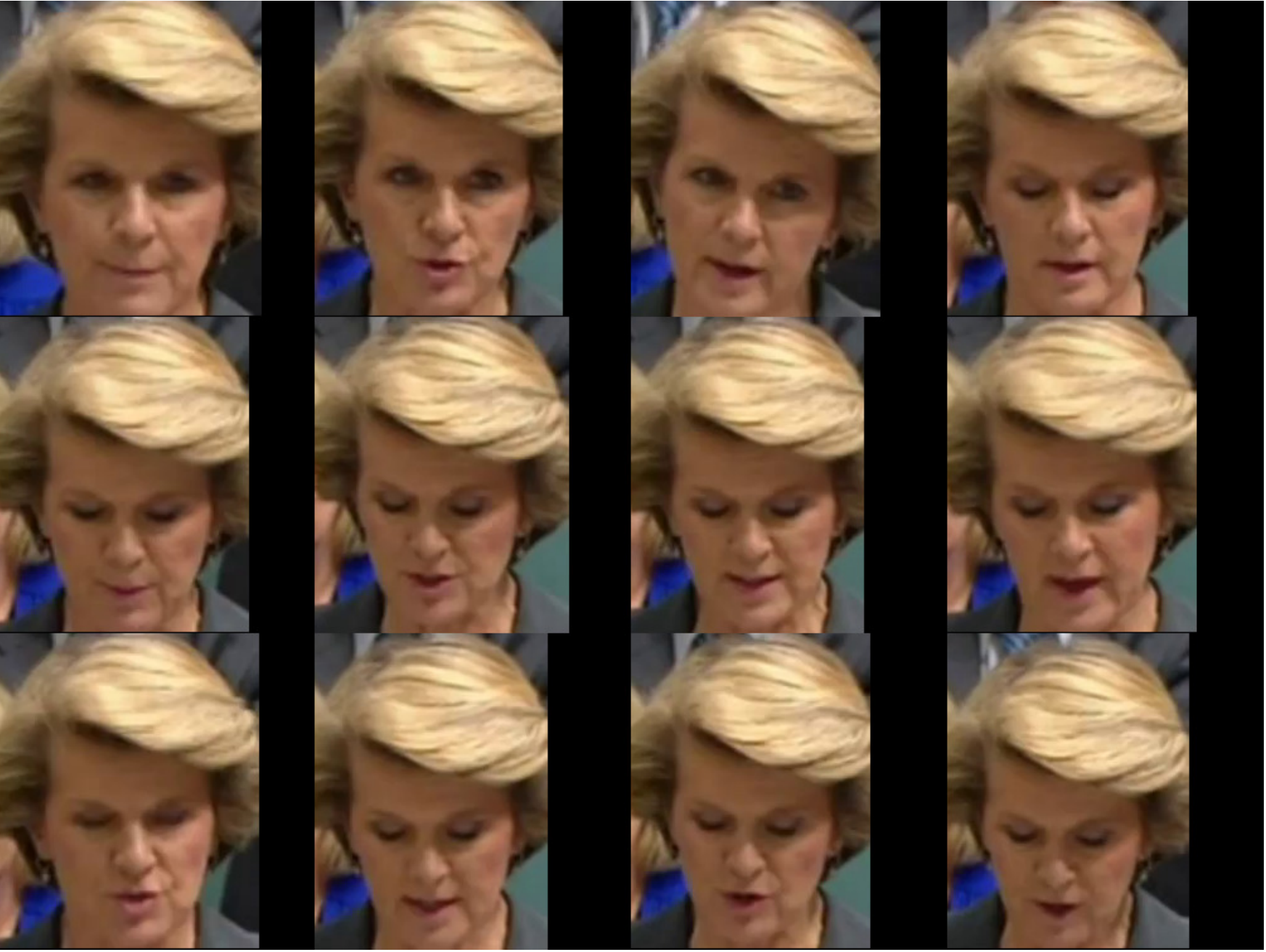}\hspace{3mm}
\includegraphics[height=0.18\linewidth]{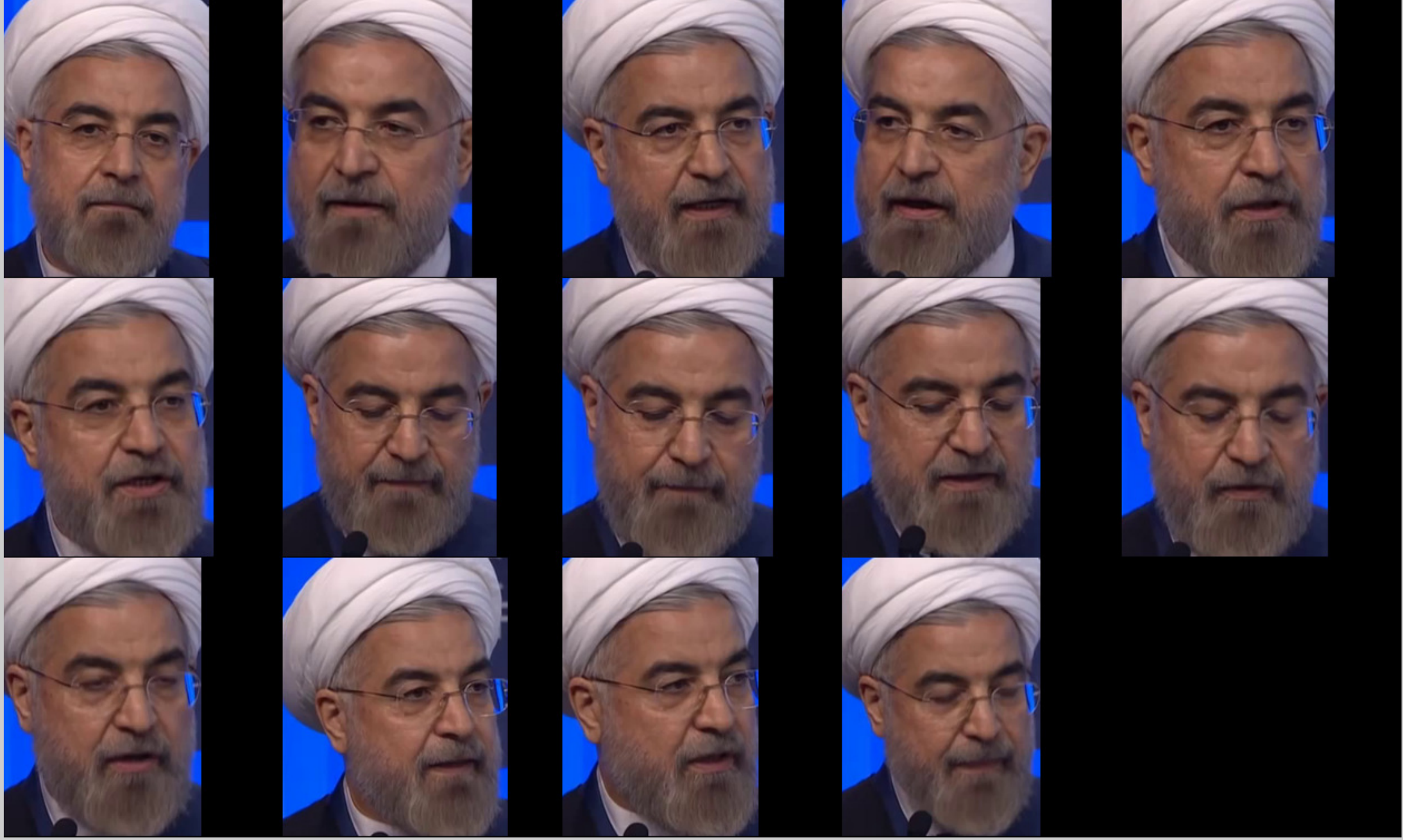}\hspace{3mm}
\includegraphics[height=0.18\linewidth]{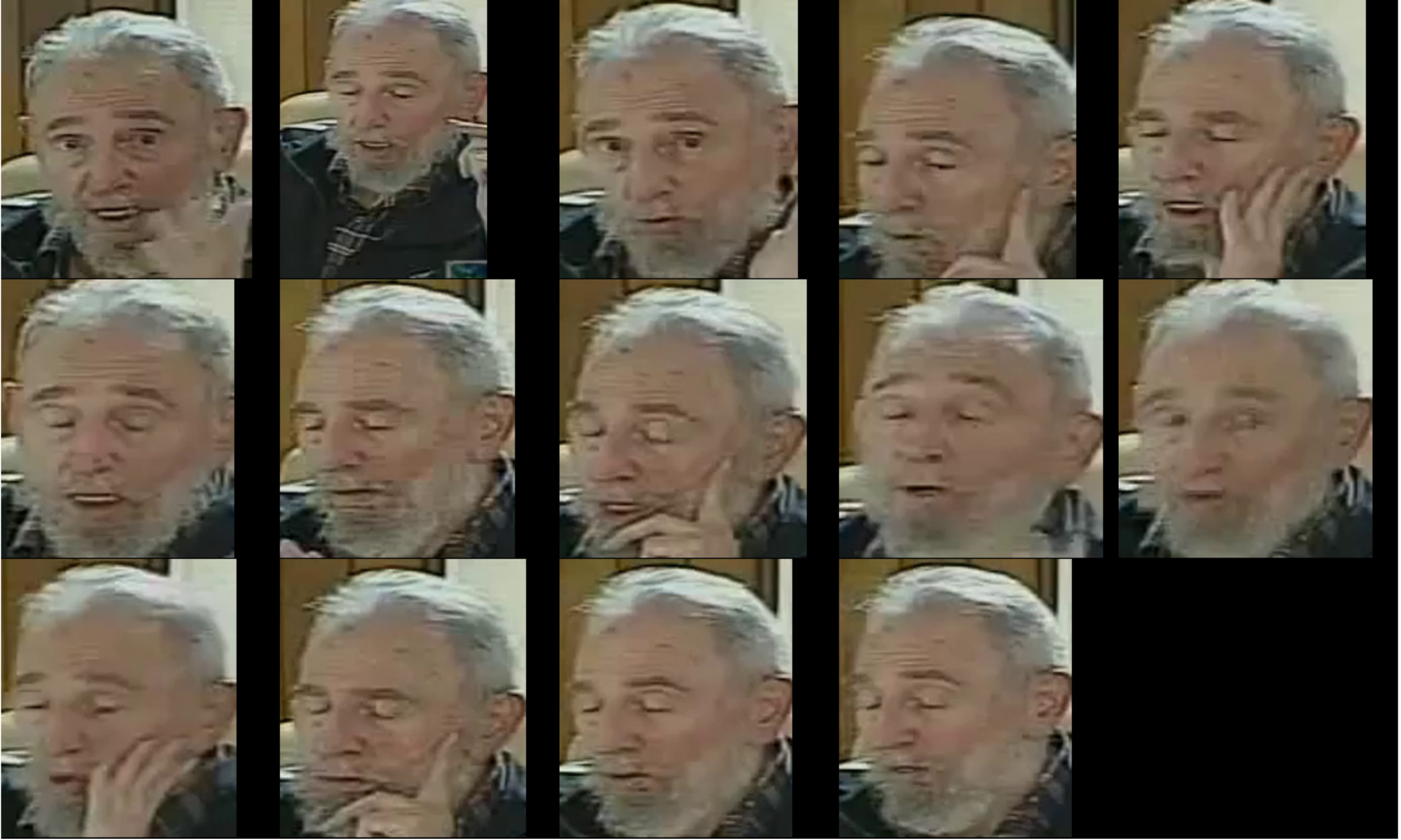}\\ 
\caption{Cropped faces from the videos of five subjects in IJB-A \cite{KlareIJBA} sorted by face image quality (best to worst) automatically predicted by the proposed HQV method (SVR on Deep-320 image features \cite{WangOtto} trained on human quality ratings from the LFW database).}
\label{fig:ijbaRankedVideos}
\end{figure*}

Figure~\ref{fig:ijbaRanked} shows examples of face images (and video frames in Fig.~\ref{fig:ijbaRankedVideos}) sorted in order of the proposed automatic face quality prediction for human quality ratings (HQV-0). Fig.~\ref{fig:ijbaRanked} also shows face images sorted by RQS \etal~\cite{Chen2015} for comparison. 
Visually, both methods appear to do reasonably well at ranking face images by quality, where both methods are noticeably sensitive to facial pose, in particular.


\begin{table}
\renewcommand{\arraystretch}{1.1}
\caption{Comparison of verification performance for the IJB-A database \cite{KlareIJBA} between previously published results and the face matchers used in this paper (with and without the proposed HQV face quality predictor). Results are reported as average $\pm$ standard deviation over the 10 folds specified by the IJB-A verify protocol.}
\centering
\small
\begin{tabular}{| R{4.7cm} |C{1.4cm}  |C{1.4cm} |}
\hline
\multicolumn{1}{|C{4.7cm}|}{\raisebox{-2mm}{Algorithm}} & \multicolumn{1}{C{1.4cm}|}{TAR @ 0.1\% FAR} & \multicolumn{1}{C{1.4cm}|}{TAR @ 1\% FAR}\\
\hline\hline
ConvNet \cite{WangOtto} (\# nets = 7) & 51.0 $\pm$ 6.1 & 72.9 $\pm$ 3.5\\
DCNN\textsubscript{\emph{all}} \cite{ChenDCNN2016} & \emph{n.a.}\textsuperscript{$\ast$} & 78.7 $\pm$ 4.3\\
DR-GAN (avg.) \cite{TranGAN2017} & 51.8 $\pm$ 6.8 & 75.5 $\pm$ 2.8\\
DR-GAN (fuse) \cite{TranGAN2017} & 53.9 $\pm$ 4.3 & 77.4 $\pm$ 2.7\\
\hline\hline
COTS-A  & 41.4 $\pm$ 3.3 & 58.0 $\pm$ 2.4\\
COTS-A w/ HQV & 61.7 $\pm$ 2.7 & 71.5 $\pm$ 1.3\\
\hline
ConvNet \cite{WangOtto} (\# nets = 1) & 30.1 $\pm$ 3.5 & 52.3 $\pm$ 3.2\\
ConvNet \cite{WangOtto} (\# nets = 1) w/ HQV & 48.0 $\pm$ 5.1 & 68.5 $\pm$ 3.5\\
\hline
\multicolumn{3}{R{6cm}}{\scriptsize \textsuperscript{$\ast$}\cite{ChenDCNN2016} did not report performance at 0.1\% FAR.}
\end{tabular}
\label{tab:ijba}
\end{table}

\section{Conclusions}
\label{sec:conclusion}

Automatic face image quality assessment is a challenging problem with important operational applications. Automatic detection of low-quality face images would be beneficial for maintaining the integrity of enrollment databases, reacquisition prompts, quality-based fusion, and adaptive recognition approaches. In this work, we 
proposed a model for automatic prediction of face image quality using image features extracted prior to matching. 
The conclusions and contributions can be summarized as follows:
\begin{itemize}
\item Human ratings of face image quality are correlated with automatic recognition performance for unconstrained face images. Rejection of $5\%$ of the lowest quality face images (based on human quality values) in the LFW database resulted in $\sim$2\% reduction in FNMR, while using human quality values to select subsets of images for template-based matching reduced FNMR by at least 13\% (at 1\% FMR) for two different matchers (COTS-A and ConvNet \cite{WangOtto}). 
\item Automatic prediction of human quality ratings (HQV) is more accurate than prediction of score-based face quality values (MQV). It is difficult to predict the score-based quality because of nuances of specific matchers and pairwise quality factors (\ie comparison scores are a function of \emph{two} face images, but we are using the scores to label the quality of a \emph{single} face image). 
\item The proposed method for face image quality performs comparably to Chen \etal's RQS \cite{Chen2015} for quality-based selection when multiple face images and videos are available for a subject. 
\item Visual inspection of face images rank-ordered by the proposed automatic face quality measures (both human ratings and score-based quality) are promising, even for cross-database prediction (\ie model trained on LFW \cite{LFWTech} and tested on IJB-A \cite{KlareIJBA} face images). 
\end{itemize}

\iftoggle{notes}{
\textcolor{blue}{Research applications: determining the ``difficulty" of a face image database. For example, the JANUS program has different stages, each of which is claimed to be more challenging than the previous. Comparisons of multiple algorithms which are evaluated on different databases are almost impossible without knowledge of the ``difficulty" (or quality) of the respective databases. Recommendations for how to summarize/aggregate the quality values of a collection of biometric samples (to reflect expected error rates) are given in \cite{NISTIR7422}.}
}

\iftoggle{notes}{
\subsection{Future Work}
\begin{itemize}
\item An entirely matcher-dependent quality measure may further distinguish face quality between three scenarios:
(i) determining face vs. non-face (flagging face detection failures), (ii) assessment of the accuracy of face alignment, and (iii) given an aligned face image, now what is the quality? 
\item \textbf{Account for the alignment/FTE issues. The lowest \eg 2\% w.r.t $z_{ij}$ may be alignment/enrollment errors for COTS, but feature extraction from deep network was successful.} 
\item The worker-rating quality matrix from matrix completion contains $n$ features per face image which could be further exploited, rather than labeling the quality of an image as simply the \emph{median} value; supervised techniques could instead be used to obtain the quality values. 
\item Classification instead of regression or a hierarchical regression, first classify into bins, then do regression in each bin for a fine-tuned ranking. 
\item Define target quality labels for only the images that are poor/good for multiple matchers (medium images could be those images where the matchers disagree). 
\item Ideally, we would compile a set of automatically extracted image features that are measurements of known quality factors that affect face recognition performance, such as pose, illumination, expression, occlusion, contrast, focus, etc. The quality algorithm would then be able to give meaningful/interpretative feedback to users or human operators. 
However, each of these quality factors are somewhat their own research area and there are no widely accepted methods for computing all of these quality measures. 
\item Experiments on SCface database (\url{http://www.scface.org/}).
\item Quality prediction for IJB-A and YTF video frames. What is the quality variation within the same video track? (Probably low variation for YTF database...)
\end{itemize}
}

The work presented here suggests the following future avenues of research for face image quality.
Face quality may need to be distinguished as three scenarios: (i) determining face vs. non-face (flagging face detection failures), (ii) assessment of the accuracy of face alignment, and (iii) given an aligned face image, now what is the quality? 
These three modules of a face image quality algorithm may allow for the integration of face matcher-dependent properties (\eg IPD, alignment errors) with more generalizable face image quality measures.
A hierarchical prediction approach may improve the overall face quality prediction accuracy. For example,
face quality of an image could first be classified as low, medium, or high (where the bins are defined to be highly correlated with recognition performance), followed by regression within each bin for a fine-tuned ranking (useful for visual purposes and
other ranking applications).
The current image features extracted from a ConvNet \cite{WangOtto} show promising results for face image quality.
However, the ConvNet \cite{WangOtto} was trained for face recognition purposes, so the representation should ideally be robust to face quality factors. It would be desirable to retrain a ConvNet for prediction of face image quality, rather than identity.

\iftoggle{notes}{
\subsection{Matcher-Dependent vs. Matcher-Independent}
Generalizability of a face image quality measure is difficult because (i) different matchers are more/less robust to different factors and (ii) different applications may require certain quality factors to take precedence over others (indoor vs. outdoor, frontal vs. unconstrained capture, etc.).
}

\iftoggle{notes}{
\begin{itemize}
\item Details about face DETECTION and ALIGNMENT.
\item LABELS vs. VALUES vs. RATINGS
\item SVR is NAIVE approach for predicting score-based values... so can we claim that predicting human quality values is more accurate?!?
\end{itemize}
}

\ifCLASSOPTIONcaptionsoff
  \newpage
\fi




\bibliographystyle{IEEEtran}
\bibliography{IEEEabrv,bare_jrnl_FaceQuality_TIFS}

%
%

%

\begin{IEEEbiography}[{\includegraphics[width=1in,height=1.25in,clip,keepaspectratio]{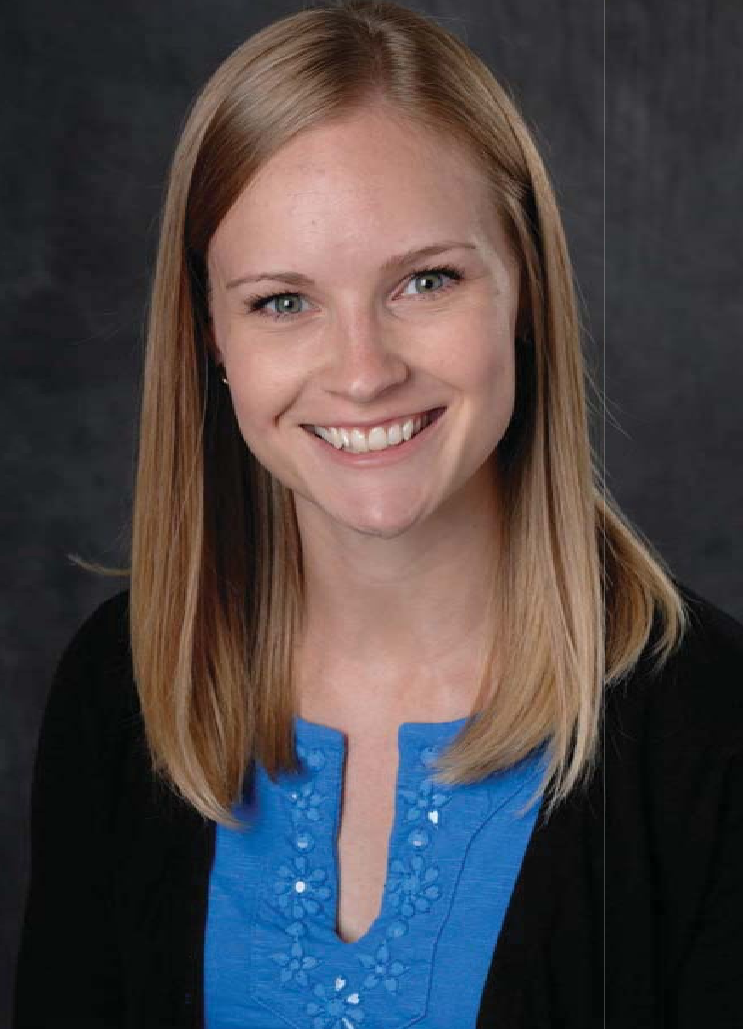}}]{Lacey Best-Rowden}
received the B.S. degree in
computer science and mathematics from Alma College,
Alma, Michigan, in 2010, and the Ph.D. from the Department
of Computer Science and Engineering at Michigan
State University, East Lansing, Michigan, in 2017. Her research
interests include pattern recognition, computer
vision, and image processing with applications
in biometrics. She is a student member of the IEEE.
\end{IEEEbiography}

\begin{IEEEbiography}[{\includegraphics[width=1in,height=1.25in,clip,keepaspectratio]{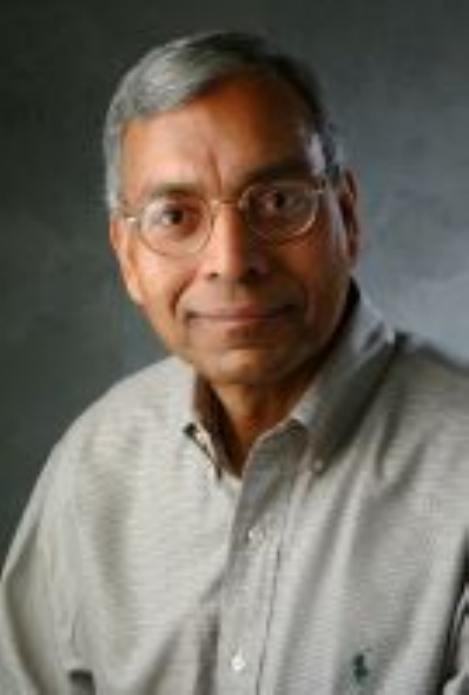}}]{Anil K. Jain}
is a University distinguished professor
in the Department of Computer Science and
Engineering at Michigan State University. He is a
Fellow of the ACM, IEEE, IAPR, AAAS and SPIE.
His research interests include pattern recognition and
biometric authentication. He served as the editor-inchief
of the IEEE Transactions on Pattern Analysis
and Machine Intelligence, a member of the United
States Defense Science Board and the Forensics
Science Standards Board. He has received Fulbright,
Guggenheim, Alexander von Humboldt, and IAPR
King Sun Fu awards. He is a member of the United States National Academy
of Engineering and a Foreign Fellow of the Indian National Academy of
Engineering.
\end{IEEEbiography}
\vfill





\end{document}